\documentclass[10pt,twocolumn,letterpaper]{article}

\usepackage{cvpr}              
\usepackage{algorithm}
\usepackage{algorithmic}
\usepackage{amsmath}
\usepackage{amssymb}
\usepackage{comment}
\usepackage{multirow}
\usepackage[accsupp]{axessibility}
\definecolor{cvprblue}{rgb}{0.21,0.49,0.74}
\usepackage[pagebackref,breaklinks,colorlinks,allcolors=cvprblue]{hyperref}

\def\paperID{35193} 
\def\confName{CVPR}
\def\confYear{2026}

\title{\textsc{IDperturb}: Enhancing Variation in Synthetic Face Generation via Angular Perturbations}
 
\author {
     Fadi Boutros\textsuperscript{\textrm 1},
    Eduarda Caldeira\textsuperscript{\textrm 1, \textrm 2},
    Tahar Chettaoui\textsuperscript{\textrm 1},
    Naser Damer\textsuperscript{\textrm 1, \textrm 2} \\
    \textsuperscript{\rm 1}Fraunhofer Institute for Computer Graphics Research IGD, Germany\\
    \textsuperscript{\rm 2} TU Darmstadt, Germany
    \\
    {\tt\small fadi.boutros@igd.fraunhofer.de}
}

\begin{document}
\maketitle
\begin{abstract}
Synthetic data has emerged as a practical alternative to authentic face datasets for training face recognition (FR) systems, especially as privacy and legal concerns increasingly restrict the use of real biometric data. Recent advances in identity-conditional diffusion models have enabled the generation of photorealistic and identity-consistent face images. However, many of these models suffer from limited intra-class variation, an essential property for training robust and generalizable FR models. In this work, we propose IDPERTURB, a simple yet effective geometric-driven sampling strategy to enhance diversity in synthetic face generation. IDPERTURB perturbs identity embeddings within a constrained angular region of the unit hyper-sphere, producing a diverse set of embeddings without modifying the underlying generative model. Each perturbed embedding serves as a conditioning vector for a pre-trained diffusion model, enabling the synthesis of visually varied yet identity-coherent face images suitable for training generalizable FR systems.
Empirical results demonstrate that training FR on datasets generated using IDPERTURB yields improved performance across multiple FR benchmarks, compared to existing synthetic data generation approaches. Code and generated datasets are publicly available \url{https://github.com/fdbtrs/IDperturb}.
\end{abstract}    
\section{Introduction}
\label{sec:intro}

Face recognition (FR) has achieved remarkable progress in recent years, driven by advances in deep neural network architectures, innovations in training objectives, particularly margin-based softmax losses, and the availability of large-scale annotated face datasets~\cite{Deng_2022,DBLP:conf/cvpr/Kim0L22,wang2018cosfacelargemargincosine}. However, acquiring diverse, ethically and legally compliant face datasets remains a significant challenge~\cite{gdpr, lirias3838501,SyntheticFRSurvay}. Increasing privacy concerns and regulatory constraints surrounding biometric data have led to the withdrawal of several major datasets such as MS-Celeb-1M~\cite{guo2016ms} and VGGFace2~\cite{DBLP:conf/fgr/CaoSXPZ18}, severely limiting access to authentic training data.

These limitations have spurred growing interest in using synthetic data as a scalable and privacy-preserving alternative for training FRs~\cite{DBLP:journals/inffus/MelziTVKRLDMFOZZYZWLTKZDBVGFFMUG24,DBLP:conf/fgr/Otroshi-Shahreza24,DEANDRESTAME2025103099}. Recent advances in generative models~\cite{Rombach2021,FFHQ,Karras2020StyleGANADA}, especially diffusion models (DMs), have enabled the synthesis of photorealistic and identity-consistent face images. Several studies have shown that identity-conditioned DMs can produce synthetic datasets that yield competitive performance compared to real data in downstream FR tasks~\cite{arc2face,DBLP:conf/iccv/BoutrosGKD23,DBLP:conf/cvpr/Kim00023,DBLP:conf/nips/Xu0WXDJHM0DH24,Caldeira_Chettaoui_Damer_Boutros_2026}. However, despite their success in maintaining identity fidelity, many of these models struggle to capture sufficient intra-class variation, an essential factor for training robust and generalizable FR systems.

In this work, we propose \textsc{IDperturb}, a simple yet effective sampling strategy for enhancing the diversity of synthetic face datasets. Our method leverages pre-trained identity-conditioned DMs~\cite{DBLP:conf/iccv/BoutrosGKD23} and perturbs identity embeddings within a constrained angular region on the unit hyper-sphere. Rather than using a fixed identity embedding~\cite{DBLP:conf/iccv/BoutrosGKD23,DBLP:conf/cvpr/Kim00023,DBLP:conf/iclr/LinHXMZD25,arc2face} for generation, \textsc{IDperturb} creates a set of perturbed embeddings per identity, each serving as a conditioning vector for a unique sample. This results in visually diverse yet semantically coherent outputs.
Unlike previous methods that rely on auxiliary labels~\cite{DBLP:conf/nips/Xu0WXDJHM0DH24}, learned style modules~\cite{DBLP:conf/cvpr/Kim00023}, or external models layered over a diffusion backbone~\cite{DBLP:conf/iclr/Otroshi-Shahreza25}, \textsc{IDperturb} is a fully geometry-driven approach that operates purely in the embedding space and is compatible with pre-trained DMs. By leveraging the geometric structure of identity embeddings, our method introduces diversity while preserving identity semantics to a large extent.
Empirically, we demonstrate that \textsc{IDperturb} significantly improves intra-class diversity and enhances downstream FR accuracy. 

\noindent \textbf{Our main contributions are:}
\begin{itemize}
    \item We propose a geometric method to perturb identity embeddings within a cosine-constrained spherical cap, enabling the generation of images with large variations of the same coherent identity.
    \item We demonstrate that  data generated with \textsc{IDperturb} yield superior performance on multiple FR benchmarks, surpassing existing state-of-the-art (SOTA) approaches.
\end{itemize}

\section{Related Work}
\label{sec:related_work}
The rapid progress of deep generative models (DGMs)~\cite{Rombach2021,FFHQ,Karras2020StyleGANADA} has enabled the synthesis of photorealistic images with high identity fidelity and visual diversity. This capability has spurred a surge of interest in using synthetic data to train FR~\cite{DEANDRESTAME2025103099,DBLP:conf/fgr/Otroshi-Shahreza24}. Existing efforts in this domain typically employ generative adversarial networks (GANs)~\cite{DBLP:conf/iccv/QiuYG00T21,Boutros2022SFace,ExFaceGAN} or DMs~\cite{DBLP:conf/cvpr/Kim00023,DBLP:conf/nips/Xu0WXDJHM0DH24} to generate training datasets for FR.

Early works  such as SynFace~\cite{DBLP:conf/iccv/QiuYG00T21} used DiscoFaceGAN~\cite{DBLP:conf/cvpr/DengYCWT20} to generate identity-conditioned images and introduced identity mix-up to increase intra-class variability. USynthFace~\cite{DBLP:conf/fgr/BoutrosKFKD23} proposed an unsupervised augmentation strategy to simulate diversity from a single synthetic instance. 
IDNet~\cite{DBLP:conf/cvpr/KolfREBKD23}, SFace~\cite{Boutros2022SFace}, and SFace2~\cite{10454585}, adopted StyleGAN2-ADA~\cite{Karras2020StyleGANADA} for class-conditional generation. ExFaceGAN~\cite{ExFaceGAN} disentangled identity attributes in GAN latent space, enabling multi-instance generation  from an unconditional model. 

DM-based approaches have recently gained attention for their stability and capacity to produce high-fidelity, identity-consistent images. IDiff-Face~\cite{DBLP:conf/iccv/BoutrosGKD23} was among the first to condition DMs on identity embeddings with contextual partial dropout applied to the embeddings to improve sample diversity. 
DCFace~\cite{DBLP:conf/cvpr/Kim00023} introduced dual conditioning with identity and style inputs to control appearance variation. ID$^3$~\cite{DBLP:conf/nips/Xu0WXDJHM0DH24} extended  IDiff-Face~\cite{DBLP:conf/iccv/BoutrosGKD23} by incorporating binary attributes as additional conditioning. Arc2Face~\cite{arc2face} fine-tuned Stable Diffusion on WebFace42M~\cite{DBLP:conf/cvpr/ZhuHDY0CZYLD021} to generate high-resolution, identity-faithful face images. HyperFace~\cite{DBLP:conf/iclr/Otroshi-Shahreza25} proposed an iterative approach to sample embedding from FR embedding space with large intra-class variations and utilized Arc2Face~\cite{arc2face} to generate synthetic data for FR training.
GANDiffFace~\cite{DBLP:conf/iccvw/MelziRTVLDS23} explored hybrid generation by combining GANs with diffusion.
The effectiveness of synthetic data for FR has been validated through recent  competitions~\cite{DBLP:journals/inffus/MelziTVKRLDMFOZZYZWLTKZDBVGFFMUG24,DBLP:conf/fgr/Otroshi-Shahreza24}. Winning solutions frequently leveraged diffusion-based synthetic datasets~\cite{DBLP:conf/iccv/BoutrosGKD23,DBLP:conf/cvpr/Kim00023}, underscoring their superiority in generating diverse and identity-separable samples.

\textbf{Identity Consistency in DMs:}
Ensuring identity preservation during generation is a key challenge in diffusion-based synthesis \cite{DBLP:conf/iccv/BoutrosGKD23,DBLP:conf/cvpr/Kim00023,DBLP:conf/nips/Xu0WXDJHM0DH24,DBLP:conf/iclr/LinHXMZD25}, especially for the downstream task of synthetic-based FR. Two main strategies have emerged to address this.
The first leverages Classifier-Free Guidance (CFG)~\cite{DBLP:conf/iclr/LinHXMZD25}, where the model is trained both with and without identity conditioning. During inference, predictions are interpolated to modulate conditioning strength. When applied to identity embeddings, CFG effectively improves identity adherence while maintaining sample realism.
The second strategy incorporates explicit identity supervision via an additional loss. For instance, DCFace \cite{DBLP:conf/cvpr/Kim00023} and ID$^3$~\cite{DBLP:conf/nips/Xu0WXDJHM0DH24} use an identity-preserving loss that minimizes the distance between the identity embedding of the generated image and the conditioning vector. This reinforces identity consistency during the denoising process. Despite the utility of both approaches, CFG-based techniques ~\cite{DBLP:conf/iclr/LinHXMZD25,arc2face} have demonstrated greater simplicity and effectiveness. Empirical results suggest that synthetic datasets generated with CFG outperform those trained with auxiliary identity losses in downstream recognition tasks~\cite{DBLP:conf/iclr/LinHXMZD25}.

\textbf{Diversity in Identity-Conditioned Generation:}
While identity conditioning ensures semantic coherence, it often results in limited intra-class variation—an undesirable property for FR training. To overcome this, 
DCFace~\cite{DBLP:conf/cvpr/Kim00023} supplements the identity condition with a learned style embedding, allowing stylistic diversity across samples. ID$^3$~\cite{DBLP:conf/nips/Xu0WXDJHM0DH24} introduces binary attributes as additional conditions, enabling explicit control of intra-class variation. 
IDiff-Face~\cite{DBLP:conf/iccv/BoutrosGKD23} applies dropout to the identity context during training to encourage generalization. HyperFace~\cite{DBLP:conf/iclr/Otroshi-Shahreza25} proposed an iterative approach to optimize intra-class variation for generating synthetic identities from a spherical space.
UIFace~\cite{DBLP:conf/iclr/LinHXMZD25} employs a two-phase sampling scheme where the model generates early steps unconditionally and incorporates the identity condition only in later denoising steps, allowing more stochastic variations.

While prior work has significantly advanced synthetic face generation for FR, balancing identity preservation and visual diversity, many approaches depend on auxiliary labels~\cite{DBLP:conf/nips/Xu0WXDJHM0DH24}, network modifications~\cite{DBLP:conf/cvpr/Kim00023}, or iterative learnable approaches for context sampling~\cite{DBLP:conf/iclr/Otroshi-Shahreza25}. In contrast, our method leverages the geometry of the identity embedding space to introduce variations via angular sampling, enhancing diversity without compromising identity fidelity. This simple strategy enables the generation of diverse samples from pre-trained DMs. 

\begin{figure}[ht!]
\centering
\includegraphics[width=\linewidth]{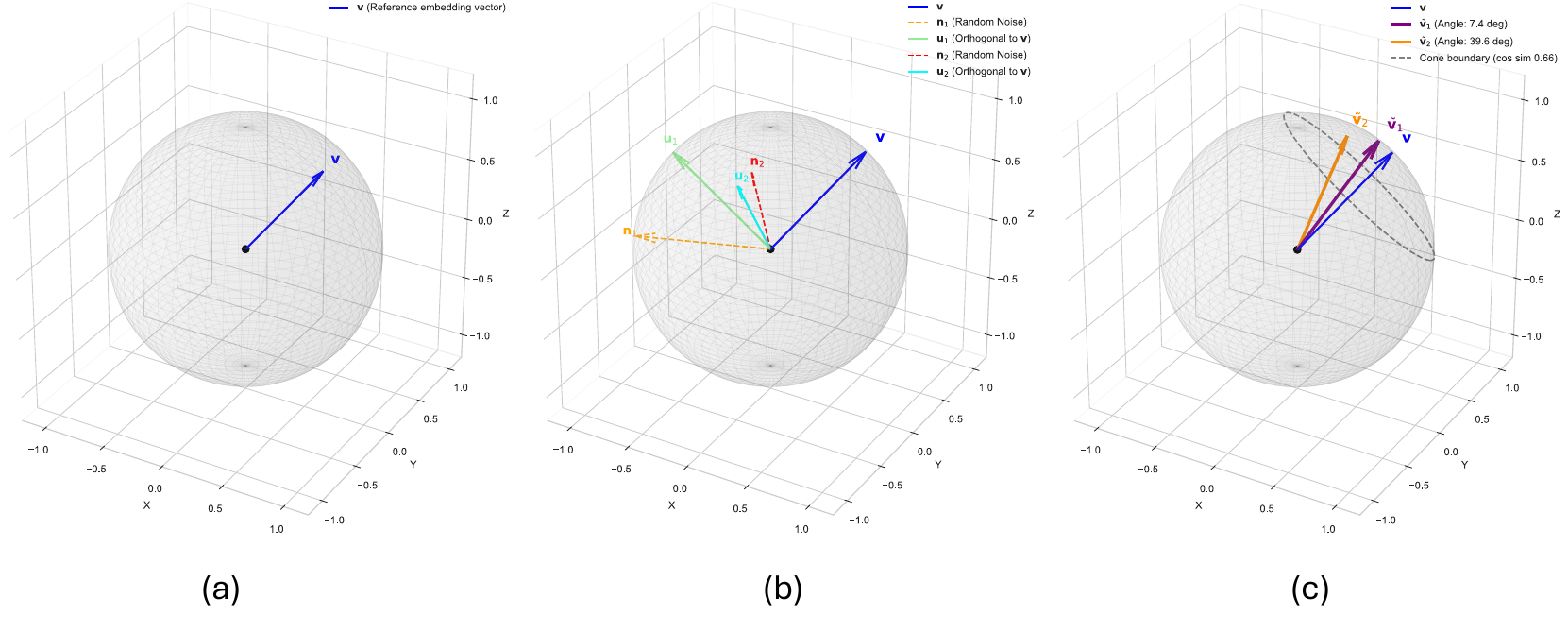} 
\vspace{-5mm}
\caption{Identity perturbation in the embedding space. (a) a normalized reference embedding \( \mathbf{v} \in \mathbb{R}^3 \). (b) random \( \mathbf{n_1}, \mathbf{n_2} \sim \mathcal{N}(0, \mathbf{I}) \) sampled and projected onto the hyperplane orthogonal to \( \mathbf{v}\), resulting in \( \mathbf{u_1}\) and \( \mathbf{u_2}\) (Equation \ref{eq:orthognal}).  (c)  perturbed identity vector \( \mathbf{\tilde{v}_1}\) and \( \mathbf{\tilde{v}_2}\) with different angle to \( \mathbf{v}\) are constructed (Equation \ref{eq:v_construct}). Note, this figure is not manually plotted, rather, using a  Python script corresponding to Algo. \ref{alg:angular_sampling_multiple}. The cone boundary, dashed circumference in (c), is defined by the lower bound \( \mathbf{lb}\) is 0.66, and \( \mathbf{n_1}\) and  \(\mathbf{n_2}\) are sampled on the fly to illustrate a real scenario. For visualization purposes, we set the number of dimensions to 3, \( \mathbf{v} \in \mathbb{R}^3 \). }
\label{fig:apporach}
\vspace{-5mm}
\end{figure}

\section{Methodology}
\label{sec:methodology}
This section presents our proposed method, \textsc{IDperturb}, a sampling strategy designed to enhance intra-class variation in synthetic face generation. Our approach operates directly in the identity-condition embedding space of a pre-trained identity-conditional DM. The core idea is to perturb the fixed identity embedding within a constrained angular region, defined as a $d$-dimensional cone, thus generating samples that are diverse yet, to a large degree, identity-consistent.
Given a pre-trained DM conditioned on identity features (i.e., embeddings extracted by a FR model), \textsc{IDperturb} perturbs the conditioning vector by controlling its angular distance from the original embedding. This process yields a set of perturbed identity embeddings, which are then used as input to the DM to synthesize face images. We first provide preliminaries on DM in this section, followed by our proposed \textsc{IDperturb}.

\subsection{Latent DM Preliminaries}
\label{sec:ldm}

Latent Diffusion Models (LDMs)~\cite{Rombach2021} improve the efficiency of diffusion-based image generation by operating in a compressed latent space rather than directly in pixel space. Given a training image \( \mathbf{x}_0 \in \mathbb{R}^{H \times W \times C} \), a pre-trained encoder \( \mathcal{E}: \mathbb{R}^{H \times W \times C} \rightarrow \mathbb{R}^{d} \) maps it to a latent representation \( \mathbf{z}_0 = \mathcal{E}(\mathbf{x}_0) \). 

To learn the generative process, LDMs adopt the denoising diffusion probabilistic model (DDPM) framework~\cite{Ho2020}, where the latent code \( \mathbf{z}_0 \) is gradually perturbed by Gaussian noise over \( T \) discrete timesteps:
\begin{equation}
q(\mathbf{z}_t | \mathbf{z}_{t-1}) = \mathcal{N}(\mathbf{z}_t; \sqrt{1 - \beta_t} \mathbf{z}_{t-1}, \beta_t \mathbf{I}),
\end{equation}
where \( \{ \beta_t \}_{t=1}^T \) is a predefined variance schedule. After sufficient noising steps, the latent \( \mathbf{z}_T \) approximates an isotropic Gaussian distribution, i.e., \( \mathbf{z}_T \sim \mathcal{N}(0, \mathbf{I}) \).

The reverse process is modeled by a neural network \( \theta \), typically implemented as a U-Net~\cite{Rombach2021}, which is trained to denoise the latent variable by predicting the noise component added at each step. The network takes as input the noisy latent \( \mathbf{z}_t \), the timestep \( t \), and an optional conditioning signal \( \mathbf{v} \in \mathbb{R}^d \), such as an identity embedding. The reverse process is thus formulated as:
\begin{equation}
\label{eq:denoise}
p_\theta(\mathbf{z}_{t-1} | \mathbf{z}_t, \mathbf{v}) = \mathcal{N}(\mathbf{z}_{t-1}; \mu_\theta(\mathbf{z}_t, t, \mathbf{v}), \Sigma_\theta(\mathbf{z}_t, t, \mathbf{v})).
\end{equation}

In identity-conditional LDMs~\cite{DBLP:conf/iccv/BoutrosGKD23,DBLP:conf/nips/Xu0WXDJHM0DH24,DBLP:conf/iclr/LinHXMZD25}, the identity embedding \( \mathbf{v} \), typically extracted using a pre-trained FR model, is injected into the U-Net via cross-attention layers~\cite{Rombach2021}. This mechanism projects the identity context into the intermediate feature representations of the denoising network, allowing the generative process to dynamically adapt to the given identity~\cite{Rombach2021,DBLP:conf/iccv/BoutrosGKD23}.
Note that the condition is dropped (i.e, condition is set to zeros) with probability $P=20$ during the training to enable unconditional sampling \cite{DBLP:conf/iclr/LinHXMZD25}.
After the denoising trajectory is complete, the final latent \( \hat{\mathbf{z}}_0 \) is passed through a pre-trained decoder \( \mathcal{D}: \mathbb{R}^{d} \rightarrow \mathbb{R}^{H \times W \times C} \), to reconstruct image \( \hat{\mathbf{x}}_0 = \mathcal{D}(\hat{\mathbf{z}}_0) \).

To generate identity-conditioned synthetic data using a pre-trained LDM, the common approach~\cite{DBLP:conf/iccv/BoutrosGKD23,DBLP:conf/nips/Xu0WXDJHM0DH24,DBLP:conf/cvpr/Kim00023,DBLP:conf/iclr/LinHXMZD25} is to first sample an image \( \mathbf{x}' \) using an unconditional generative process (i.e., conditioning input of zeros). A feature representation \( \mathbf{v}' = f(\mathbf{x}') \) is then extracted using a pre-trained FR model \( f \). This embedding \( \mathbf{v}' \) serves as the identity condition to generate new samples from the conditional LDM by fixing \( \mathbf{v}' \) and varying the initial noise seed, \(\mathbf{z}_T\)~\cite{DBLP:conf/iccv/BoutrosGKD23,DBLP:conf/iclr/LinHXMZD25}.

\textbf{Classifier-Free Guidance (CFG):}
CFG~\cite{DBLP:journals/corr/abs-2207-12598} can be utilized to amplify the effect of the identity condition during generation \cite{DBLP:conf/iclr/LinHXMZD25}. CFG has proven effective across a range of conditional generation tasks, including both image- and text-guided scenarios~\cite{ban2024understanding, DBLP:conf/icml/Wang0HG24, DBLP:conf/iclr/LinHXMZD25}.
At inference time, the predicted noise is interpolated between the conditioned and unconditioned outputs as follows:
\begin{equation}
\label{eq:cfg}
\hat{\epsilon} = (1 + \omega)\hat{\epsilon}_\theta(\mathbf{z}_t, t, \mathbf{v}') - \omega \hat{\epsilon}_\theta(\mathbf{z}_t, t),
\end{equation}
where \( \omega \) is the guidance strength parameter. Increasing \( \omega \) enhances the model's adherence to the identity condition, promoting better identity consistency in the generated data.

\subsection{Data Generation via Angular Sampling}

For identity-conditioned generation, particularly in applications such as facial synthesis for training FRs, the generated data must simultaneously fulfill two critical objectives: maintaining high identity consistency across samples of the same identity and exhibiting realistic intra-identity variation. These properties are paramount for the effective utilization of synthetic data in downstream FR tasks.
Existing methodologies often rely on auxiliary modules~\cite{DBLP:conf/cvpr/Kim00023,DBLP:conf/nips/Xu0WXDJHM0DH24} or employ complex learned sampling strategies on top of generative models~\cite{DBLP:conf/iclr/Otroshi-Shahreza25} to introduce diversity. In contrast, we introduce Angular Sampling, a novel geometric approach designed to stochastically perturb the conditioning identity vector in a semantically coherent manner, without requiring architectural modifications of identity-conditional DMs. 

\textbf{Assumption} We assume that the pre-trained DM, conditioned on identity embeddings, generates, to a large extent, identity-consistent samples, a property we empirically verify in our experiments.
There is no additional condition, e.g., binary attribute conditions or style conditions from a style bank. This assumption is essential to disentangle the source of intra-class variations in this work.

\textbf{Angular Sampling Formulation}
The key idea of our approach is to perturb the identity condition \( \mathbf{v} \) geometrically within the identity embedding space. This preserves, to a large degree, semantic identity while introducing diversity without modifying the base LDM. 

Let \( \mathbf{v} \in \mathbb{R}^d \) be a unit-norm identity embedding extracted from an FR model. Our goal is to sample a perturbed embedding \( \tilde{\mathbf{v}} \) such that the angular separation, \( \angle(\mathbf{v}, \tilde{\mathbf{v}}) \), is constrained within a $d$-dimensional cone defined by cosine similarity bounds \( [\mathbf{lb}, 1] \subset [0, 1] \), where $\mathbf{lb}$ is the lower bound. The procedure for generating \( \tilde{\mathbf{v}} \) for each synthetic image is illustrated in Figure~\ref{fig:apporach} and is defined as follows.
To generate a set of images corresponding  to  identity condition \( \mathbf{v}\), we first uniformly sample a target cosine similarity \( s \sim \mathcal{U}[\mathbf{lb}, 1] \), corresponding to angle \( \theta = \cos^{-1}(s) \). Then, we sample random noise \( \mathbf{n} \sim \mathcal{N}(0, \mathbf{I}) \) where \(\mathbf{n} \in \mathbb{R}^d \). We then project \(\mathbf{n}\) onto the hyperplane orthogonal to  \( \mathbf{v}\) by subtracting the component of \(\mathbf{n}\) in the direction of \( \mathbf{v}\): 
\begin{equation}
    \label{eq:orthognal}
        \mathbf{u} = \frac{\mathbf{n} - (\mathbf{n} \cdot \mathbf{v}) \mathbf{v}}{\| \mathbf{n} - (\mathbf{n} \cdot \mathbf{v}) \mathbf{v} \|}.
\end{equation}
This projection removes the contribution of \( \mathbf{v}\) from \(\mathbf{n}\), ensuring that \(\mathbf{u}\) is orthogonal to \( \mathbf{v}\). The normalization step guarantees that \(\mathbf{u}\)  has unit norm. Geometrically, \(\mathbf{u}\)  lies on the unit tangent hyper-sphere at \( \mathbf{v} \), and together with \( \mathbf{v} \), they span a \(\mathbb{R}^d \) subspace in which angular sampling is performed.
Finally, we construct the perturbed identity:
\begin{equation}
    \label{eq:v_construct}
        \tilde{\mathbf{v}} = \cos(\theta) \cdot \mathbf{v} + \sin(\theta) \cdot \mathbf{u},
    \end{equation}
where $\theta$ is the target angle sampled via $\theta = \cos^{-1}(s)$, with $s \sim \mathcal{U}[\mathbf{lb}, 1]$. Since $v$ and $u$ are orthonormal, $\|\tilde{v}\| = 1$ and $\langle \tilde{\mathbf{v}}, \mathbf{v} \rangle = \cos(\theta)$ hold, providing strict control over the perturbation magnitude while preserving identity semantics.

\textbf{Properties} The constructed perturbed embedding \( \tilde{\mathbf{v}} \) satisfies the following geometric and semantic guarantees: 
\begin{itemize}[noitemsep,nolistsep]
   \item  \textbf{Norm Preservation:} \( \| \tilde{\mathbf{v}} \|^2 = \cos^2(\theta) + \sin^2(\theta) = 1 \)
   \item  \textbf{Angle Control:} \( \langle \tilde{\mathbf{v}}, \mathbf{v} \rangle = \cos(\theta) = s \)
  \item \textbf{$\theta = 0$}: \(\tilde{\mathbf{v}}\) and \(\mathbf{v}\) are identical.
\textbf{Proof}:  \( \| \tilde{\mathbf{v}} \|= \| \mathbf{v}  \| =1 \),   \(\cos(\theta) =1\) , \(\sin(\theta) =0\), \( \tilde{\mathbf{v}}= 1 \cdot \mathbf{v} + 0 \cdot \mathbf{u},\)
\end{itemize}

\textbf{Lower Bound Constraint}
The parameter \( \mathbf{lb} \) (lower bound) delineates the permissible range for the cosine similarity between the original identity embedding \( \mathbf{v} \) and its perturbed counterpart \( \tilde{\mathbf{v}} \). Geometrically, this defines a spherical cap around \( \mathbf{v} \). A smaller value of \( \mathbf{lb} \) facilitates greater variation in the generated samples, whereas values approaching $1$ enforce stricter identity faithfulness. Notably, when \( \mathbf{lb} = 1 \), \( \mathbf{v} \) and \( \tilde{\mathbf{v}} \) are identical, resulting in no perturbation.

\textbf{Avoiding Identity Overlap}
Consider a set of reference synthetic identity embedding $\mathbf{V}$ extracted  using a pre-trained FR model from a set of reference images $\mathbf{X}$, i.e., synthetic images generated using an unconditional generative model \cite{DBLP:conf/iccv/BoutrosGKD23,DBLP:conf/iclr/LinHXMZD25}.
To mitigate the risk of the perturbed vector \( \tilde{\mathbf{v}} \) from \( \mathbf{v_i} \in  \mathbf{V}\)  becoming semantically closer to any other distinct identities, \( \mathbf{v_j} \in  \mathbf{V}\), than to its intended identity \( \mathbf{v_i} \), we dynamically adjust the lower bound \( \mathbf{lb} \):
\begin{equation}
   \mathbf{lb} \leftarrow \max\left(\mathbf{lb}, \max_{j \ne i} \text{cos} \left(\frac{\angle(\mathbf{v_i}, \mathbf{v_j})}{2} \right)\right),
\end{equation}
where dividing the angle that separates \( \mathbf{v_i} \) and \( \mathbf{v_j} \) in half ensures that  \( \tilde{\mathbf{v}} \) lays closer to \( \mathbf{v_i} \) than \( \mathbf{v_j} \) (\(\angle(\mathbf{\tilde{v}}, \mathbf{v_i}) \leq \angle(\mathbf{\tilde{v}}, \mathbf{v_j})\)).

\textbf{Data Generation}
For each synthetic identity embedding $\mathbf{v} \in \mathbf{V}$, we apply the angular sampling method described in Algorithm~\ref{alg:angular_sampling_multiple} to generate a set of $K$ perturbed identity embeddings $\{\tilde{\mathbf{v}}_k\}_{k=1}^K$.
These perturbed embeddings serve as identity-conditioning inputs to a pre-trained LDM. Specifically, for each $\tilde{\mathbf{v}}_k$, we sample a noise latent $\mathbf{z}_{T_k} \sim \mathcal{N}(0, \mathbf{I})$ and apply the reverse diffusion process (Equation \ref{eq:denoise}): \( \mathbf{z}_{0_k} = \text{LDM}(\mathbf{z}_{T_k}, \tilde{\mathbf{v}}_k),\)
where the identity condition $\tilde{\mathbf{v}}_k$ remains fixed across all timesteps $t = T, \ldots, 1$. The final synthetic image $\hat{\mathbf{x}}_k$ is obtained by decoding the denoised latent using the LDM decoder $D$: $\hat{\mathbf{x}}_k = D(\mathbf{z}_{0_k})$. By repeating this process for multiple $\tilde{\mathbf{v}}_k$, we construct a set of diverse yet identity-coherent face images $\{\hat{\mathbf{x}}_k\}_{k=1}^K$ associated with a single synthetic identity $\mathbf{v}$. This set can then be used as training data for FR models.

\begin{algorithm}[H]
\caption{Angular Sampling Algorithm}
\small
\label{alg:angular_sampling_multiple}
\begin{algorithmic}[1]
\REQUIRE Reference identity embedding $\mathbf{v_i} \in \mathbf{V}$, lower bound $\mathbf{lb} \in [0, 1]$,  number of samples per identity $K$
\ENSURE Set of perturbed identity vectors $\{\mathbf{\tilde{v}_{i_k}}\}_{k=1}^{K}$
\STATE Normalize $\mathbf{v_i} \leftarrow \frac{\mathbf{v_i}}{\|\mathbf{v_i}\|}$
\STATE $\mathbf{lb}  \leftarrow \max\left(\mathbf{lb}, \max_{j \ne i} \text{cos} \left(\frac{\angle(\mathbf{v_i}, \mathbf{v_j})}{2} \right)\right), \mathbf{v_j} \in \mathbf{V}$
\FOR{$k = 1$ to $K$}
    \STATE Sample $s \sim \mathcal{U}[\mathbf{lb},1]$
    \STATE Compute $\theta \leftarrow \cos^{-1}(s)$
    \STATE Sample noise $\mathbf{n} \sim \mathcal{N}(0, \mathbf{I}_d)$
    \STATE Project: $\mathbf{u} \leftarrow \frac{\mathbf{n} - (\mathbf{n} \cdot \mathbf{v_i}) \mathbf{v_i}}{\| \mathbf{n} - (\mathbf{n} \cdot \mathbf{v_i}) \mathbf{v_i} \|}$
    \STATE Compute perturbed vector: $\mathbf {\tilde{v}_{i_k}} \leftarrow \cos(\theta)\mathbf{v_i} + \sin(\theta)\mathbf{u}$
\ENDFOR
\RETURN $\{\mathbf{\tilde{v}_{i_k}}\}_{k=1}^{K}$
\end{algorithmic}
\end{algorithm}
\vspace{-2mm}

\begin{figure}[ht!]
\includegraphics[width=0.85\linewidth]{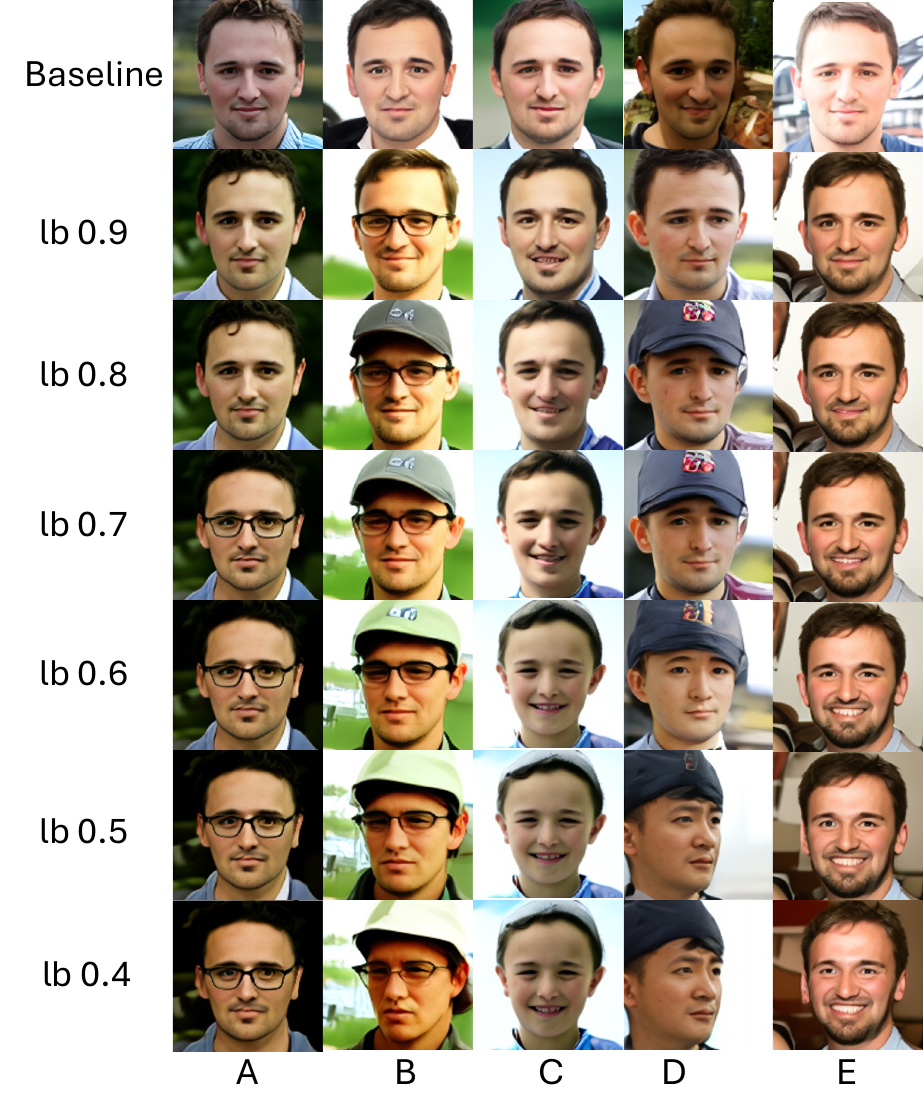} 
\vspace{-4mm}
\caption{
Effect of \textsc{IDperturb} (via lower bound \( \mathbf{lb} \)) on image diversity. The first row varies only the noise seed with fixed condition $\mathbf{v}$, subsequent rows use perturbed identity embeddings $\mathbf{\tilde{v}}$  from $\mathbf{v}$  with decreasing \( \mathbf{lb} \) (0.9 to 0.4). Lower \( \mathbf{lb} \) values increase angular deviation, enhancing intra-class variation. Note that using low lb of e.g., 0.4, the identity of some samples could be less consistent, which is also quantified by high EER in Table \ref{tab:ablation_lb}.
}
\label{fig:sample}
\vspace{-3mm}
\end{figure}

\begin{table*}[ht!]
\centering
\caption{
Identity separability of datasets (left side) and verification accuracy (\%)  of FR models trained on 0.5m images of each of those datasets  (right side).
The data was generated by varying the $d$-dimensional cone boundary $\mathbf{lb} \in [0.4,0.9]$  on two baseline LDM models: IDiff-Face trained on FFHQ and IDiff-Face trained on C-WF. 
The FR results demonstrated that \textsc{IDperturb} achieves consistent and superior performance over the baselines across all FR benchmarks. The results of the first row were obtained from authentic C-WF. 
}
\vspace{-2mm}
\resizebox{0.95\linewidth}{!}{%
\begin{tabular}{cc|ccccccccc||cccccc}
\hline
                                        &   & \multicolumn{8}{c}{\textbf{Identity-Separability of Training Data}}    & \phantom{abc}& \multicolumn{6}{c}{\textbf{FR Verification}}    \\
\hline
                                        &   & \multicolumn{2}{c}{\textbf{\multirow{2}{*}{Operation Metrics}}} & \phantom{abc} & \multicolumn{5}{c}{\textbf{Score Distributions}} & \phantom{abc}    & \multicolumn{6}{c}{\multirow{2}{*}{\textbf{Verification Benchmarks}}}    \\                              
  &     &&&&  \multicolumn{2}{c}{\textbf{genuine}} &   \multicolumn{2}{c}{\textbf{imposter}} &  \multicolumn{1}{c}{}&  \phantom{abc} &&&  &  \phantom{abc} & &  \\

\multicolumn{1}{c}{\textbf{Data Sampling Method}} & \textbf{$\mathbf{lb}$} & EER $\downarrow$  & FMR100 $\downarrow$  && G-mean & G-std & I-mean & I-std & FDR $\uparrow$  & & LFW       $\uparrow$                & AgeDB-30        $\uparrow$          & CFP-FP $\uparrow$                    & CALFW   $\uparrow$                  & CP-LFW  $\uparrow$                  & Average   $\uparrow$                       \\ \hline
\multicolumn{1}{c}{C-WF \cite{DBLP:journals/corr/YiLLL14a,DBLP:conf/iccv/BoutrosGKD23} }& - & 0.076   & 0.092 & \phantom{abc} &  0.536 &  0.215  &  0.003  &  0.070  &  5.541 & \phantom{abc}  & 99.55 & 94.55 & 95.31 & 93.78 & 89.95 & 94.63                       \\ \specialrule{.2em}{0em}{0em} 
\multicolumn{1}{c}{Baseline (FFHQ)}& - & 0.005 & 0.004 & \phantom{abc} & 0.509 & 0.104 & 0.023 & 0.081  & 13.680 & \phantom{abc}   & 97.60                     & 84.10                     & 83.36                     & 89.08                     & 78.77                     & 86.58                        \\ \hline
\multirow{6}{*}{\textsc{IDperturb}}     & 0.9  & 0.010 & 0.011 & \phantom{abc} & 0.448 & 0.108 & 0.023 & 0.060 & 11.817 & \phantom{abc}  & 98.05                     & 86.78                     & 83.79                     & 89.92                     & 79.02                     & 87.51      \\
                                        & 0.8  & 0.023 & 	0.038 & \phantom{abc} & 0.390				 & 0.116 & 0.023 & 0.060 & 7.961 & \phantom{abc}    & 98.22                     & 87.42                     & 84.26                     & 90.65                     & 81.20                     & 88.35  \\
                                        & 0.7  & 0.047 & 0.102 & \phantom{abc} & 0.337 & 0.124 & 0.022 & 0.060 & 5.236 & \phantom{abc}  & 98.47                     & 88.25                     & 84.50                     & 91.18                     & 81.27                     & 88.73   \\
                                        & 0.6 & 0.082 & 0.208 & \phantom{abc} & 0.288 & 0.130 & 0.023 & 0.060 & 3.422  & \phantom{abc}  & 98.43                    & 88.77                     & 83.61                     & 91.07                     & 80.80                     & 88.54   \\
                                        & 0.5  & 0.124 & 	0.332 & \phantom{abc} & 0.245 & 0.133 & 0.023 & 0.060 &   2.316 & \phantom{abc}   & 98.55	& 88.85	& 84.27	& 91.42 & 	80.85	& \textbf{88.79}          \\
                                        & 0.4  & 0.171 & 0.457 & \phantom{abc} & 0.208 & 0.134
                                        & 0.023 &	0.060 & 1.588 & \phantom{abc}   & 98.48	& 88.70	& 83.00	& 91.57	& 76.13	& 87.58    \\ \hline \hline

\multicolumn{1}{c}{Baseline (C-WF)} & - & 0.003 & 0.001 & \phantom{abc} & 0.670 & 0.107 &	0.010 & 0.060 & 29.116 & \phantom{abc} & 98.75 & 88.85 & 91.61 & 90.90 & 86.15 & 91.25  	\\ \hline
\multirow{6}{*}{\textsc{IDperturb}}                    & 0.9  & 0.006 & 0.005 & \phantom{abc} & 0.580 & 0.121 & 0.010 & 0.061 & 17.728 & \phantom{abc}  & 99.13 & 91.30 & 93.00 & 92.18 & 87.80 & 92.68   \\
                                        & 0.8   & 0.014 &	0.016 & \phantom{abc} & 0.494 &	
                                        0.138 & 0.009 & 0.059 & 10.471 & \phantom{abc} & 99.30 & 92.13 & 93.73 & 93.07 & 88.30 & 93.31       \\
                                        & 0.7   & 0.032 & 0.053 & \phantom{abc} & 0.412 &
                                        0.153 & 0.008 & 0.059 & 6.083 & \phantom{abc} & 99.28 & 92.37 & 93.80 & 93.38 & 88.37 & 93.44   \\
                                        & 0.6  & 0.063 & 0.134 & \phantom{abc} &  0.338 & 0.162 & 0.008 &
                                        0.059 & 3.664 & \phantom{abc}  & 99.40 & 93.20 & 93.61 & 93.50 & 88.37 & \textbf{93.62}   \\
                                        & 0.5 & 0.104 & 0.253 & \phantom{abc} & 0.278 & 0.166
                                         & 0.008 &	 0.059& 2.363 & \phantom{abc}  & 99.27 & 93.25 & 93.67 & 93.58 & 88.02 & 93.56        \\ 
                                        & 0.4  &
                                      0.152 &	
                                      0.254  & \phantom{abc} & 0.228 & 0.164 &
                                      0.008 &	0.059 & 1.599 & \phantom{abc}  & 99.28 & 93.25 & 93.06 & 93.62 & 87.62 & 93.36   \\ \hline
\end{tabular}}
\label{tab:ablation_lb}
\vspace{-2mm}
\end{table*}

\vspace{-2mm}
\section{Experimental Setup}
\label{sec:experimental_setups}
\vspace{-1mm}

\textbf{Baseline DMs} We evaluate our approach on top of two instances of pre-trained IDiff-Face \cite{DBLP:conf/iccv/BoutrosGKD23} with a contextual partial dropout of 25\% as the base DM. IDiff-Face is a conditional LDM trained on the latent space of a pre-trained Auto-Encoder \cite{Rombach2021} and conditioned on identity embeddings from a pre-trained FR. 
The first instance is trained on the Flickr-Faces-HQ (FFHQ) dataset \cite{FFHQ}, following \cite{DBLP:conf/iccv/BoutrosGKD23,DBLP:conf/nips/SunSPT24}, and it was publicly released\footnote{\url{https://github.com/fdbtrs/IDiff-Face}} by \cite{DBLP:conf/iccv/BoutrosGKD23}. The second instance of IDiff-Face \cite{DBLP:conf/iccv/BoutrosGKD23} was trained on Casia-WebFace (C-WF), and the pre-trained model was publicly released\footnote{\label{fn:tencent}\url{https://github.com/Tencent/TFace/}} by \cite{DBLP:conf/iclr/LinHXMZD25}. The choice of model trained on FFHQ \cite{DBLP:conf/iccv/BoutrosGKD23,DBLP:conf/nips/SunSPT24}, and C-WF \cite{DBLP:conf/iclr/LinHXMZD25,DBLP:conf/cvpr/Kim0L22} is to provide comparable results with SOTA.

\textbf{Synthetic data generation}
We utilized synthetic reference embeddings publicly released\footnotemark[\getrefnumber{fn:tencent}] by UIFace \cite{DBLP:conf/iclr/LinHXMZD25}. It is based on generating a set of reference images using  unconditional IDiff-Face \cite{DBLP:conf/iccv/BoutrosGKD23}.
Then, feature representations were extracted using a pre-trained FR \cite{ElasticFace}. 
Subsequently, we utilized our  \textsc{IDperturb} described in Algo.\ \ref{alg:angular_sampling_multiple} to generate 50 images per identity.
We generated 20 datasets using IDiff-Face \cite{DBLP:conf/iccv/BoutrosGKD23}, each with different \(\mathbf{lb}\) of 0.9, 0.8, 0.7, 0.6, 0.5 and 0,4 as well as using different baseline DMs and $\omega$. In the inference phase, we utilized a Denoising Diffusion Implicit Model (DDIM) \cite{DBLP:conf/iclr/SongME21} with 50 steps, following \cite{DBLP:conf/iclr/LinHXMZD25} with a fixed random seed of 1337, ensuring consistent generation across different runs. \textsc{IDperturb} induces negligible computation overhead to DM sampling with an extra 0.01 seconds to perturb 1 identity 50 times on  M3 CPU processor.

\textbf{Identity-Separability Evaluation}
We evaluate the identity-separability of synthetic data generated with \textsc{IDperturb} and compare it to the case where data is generated without the identity perturbation. 
These evaluations are reported as FMR100, which is the lowest false non-match rate (FNMR) for a false match rate (FMR)$\leq$1.0\%, along with the Equal Error Rate (EER), as in \cite{DBLP:conf/iccv/BoutrosGKD23,Boutros2022SFace}.
The inter-class separability and intra-class compactness are evaluated by reporting the genuine and imposter means (G-mean and I-mean) and standard deviation (G-STD and I-STD), respectively. Additionally, we report the Fisher discriminant ratio (FDR) \cite{poh2004study} to quantify the separability of genuine and impostor scores \cite{DBLP:conf/iccv/BoutrosGKD23,Boutros2022SFace}. 
We used a pre-trained ResNet100 \cite{He2015DeepRL} with ElasticFace \cite{ElasticFace} on MS1MV2 \cite{Deng_2022,guo2016ms} to extract the feature representations needed for these evaluations.

\textbf{Intra-Class Diversity and Consistency}
To quantify intra-class diversity induced by  \textsc{IDperturb}, we first analyze three complementary metrics, age deviation, facial expression variation, and head-pose variation, across images generated under different cone boundaries. 
Age variation was estimated using a pre-trained age estimation model \cite{karkkainenfairface}, measuring the entropy of predicted ages within each identity. 
Facial expression variation was estimated using a pretrained estimator \cite{SMW20}, measuring the entropy of predicted expression within each identity.
Head-pose diversity was assessed using a pretrained head-pose regressor \cite{Ruiz_2018_CVPR_Workshops} by computing the standard deviation (STD) of yaw–pitch–roll angles per identity. Intra-Class diversity was measured using the Learned Perceptual Image Patch Similarity (LPIPS) \cite{zhang2018perceptual} computed between intra-identity image pairs to quantify perceptual variation in appearance:
\begin{equation}
\begin{aligned}
D_{intra} &= \frac{1}{C} \sum_{i=1}^{C} 
\frac{2}{M_i(M_i-1)}
\sum_{j=1}^{M_i-1}
\sum_{l=j+1}^{M_i} \\
&\quad \left(\text{LPIPS}(x_j^{(i)},x_l^{(i)})\right),
\end{aligned}
\end{equation}
where \(C\) is number of classes, \(M_i\) is number of samples per class $i$ and LPIPS \cite{zhang2018perceptual} is perceptual similarity metric.

The intra-class consistency ($C_{intra}$) is reported by measuring the ratio of feature $f(x_j^{(i)})$ of sample $j$ of class $i$ being close to its class center $\mu_i$. $C_{intra}$ is defined, following \cite{DBLP:conf/cvpr/Kim0L22}, as:
\vspace{-1mm}
\begin{equation}
    C_{intra}= \frac{1}{C} \sum_{i=1}^{C} \frac{1}{M_i} 
\sum_{j=1}^{M_i}  \mathbb{I} \left(\text{sim}\big(f(x_j^{(i)}), \mu_i\big) \ge r \right),
\end{equation}
where the class center
$ \mu_i = \frac{1}{M_i} \sum_{j=1}^{M_i} f(x_j^{(i)})$, \(C\) is number of classes, \(M_i\) is number of samples per class $i$,
\(f(x_j^{(i)})\) is feature embedding of the \(j\)-th sample in class \(i\) extracted from \cite{ElasticFace}, \(\text{sim}(\cdot, \cdot)\) is  cosine similarity,  \(r\) is similarity threshold of value 0.3 \cite{DBLP:conf/cvpr/Kim0L22} and $\mathbb{I}(\cdot)$ is the indicator function that returns $1$ if the condition is true 
and $0$ otherwise.

\textbf{FR Training Setup}
We utilized ResNet50 \cite{He2015DeepRL} as a network architecture with CosFace loss \cite{wang2018cosfacelargemargincosine} to train FR on our synthetic datasets, following \cite{DBLP:conf/iccv/BoutrosGKD23,ExFaceGAN,DBLP:conf/iclr/LinHXMZD25,DBLP:conf/nips/SunSPT24}.  We set the mini-batch size to 512 and train with a Stochastic Gradient Descent (SGD) optimizer for 34 epochs, setting the momentum to 0.9 and the weight decay to 5e-4, following \cite{DBLP:conf/iccv/BoutrosGKD23,ExFaceGAN,DBLP:conf/nips/SunSPT24}. Initial learning rate is set to $0.1$ and is reduced by a factor of 0.1 at epochs 22, 28 and 32.
The margin penalty of CosFace loss was set to 0.35 and the scale factor to 64 \cite{wang2018cosfacelargemargincosine}.

\textbf{Evaluation Benchmarks}
We report FR evaluation results, following  \cite{DBLP:conf/iccv/BoutrosGKD23,ExFaceGAN,DBLP:conf/iclr/LinHXMZD25,DBLP:conf/cvpr/Kim0L22}, as the verification accuracy on five benchmarks, Labeled Faces in the Wild LFW \cite{LFWDatabase}, AgeDb-30 \cite{AgeDB30Database}, Cross-Age LFW (CA-LFW) \cite{CALFWDatabase}, Celebrities in Frontal-Profile in the Wild (CFP-FP) \cite{CFPFPDatabase}, and Cross-Pose LFW (CP-LFW) \cite{CPLFWDatabase}, following their official evaluation protocol. 
Additionally, we evaluated on the large-scale IARPA Janus Benchmark–C (IJB-C) \cite{DBLP:conf/icb/MazeADKMO0NACG18}. We used the official 1:1 mixed verification protocol and reported the verification performance as True Acceptance Rates (TAR) at False Acceptance Rates (FAR) of 1e-4 and 1e-5.

\begin{figure}[ht!]
\centering
\includegraphics[width=0.9\linewidth]{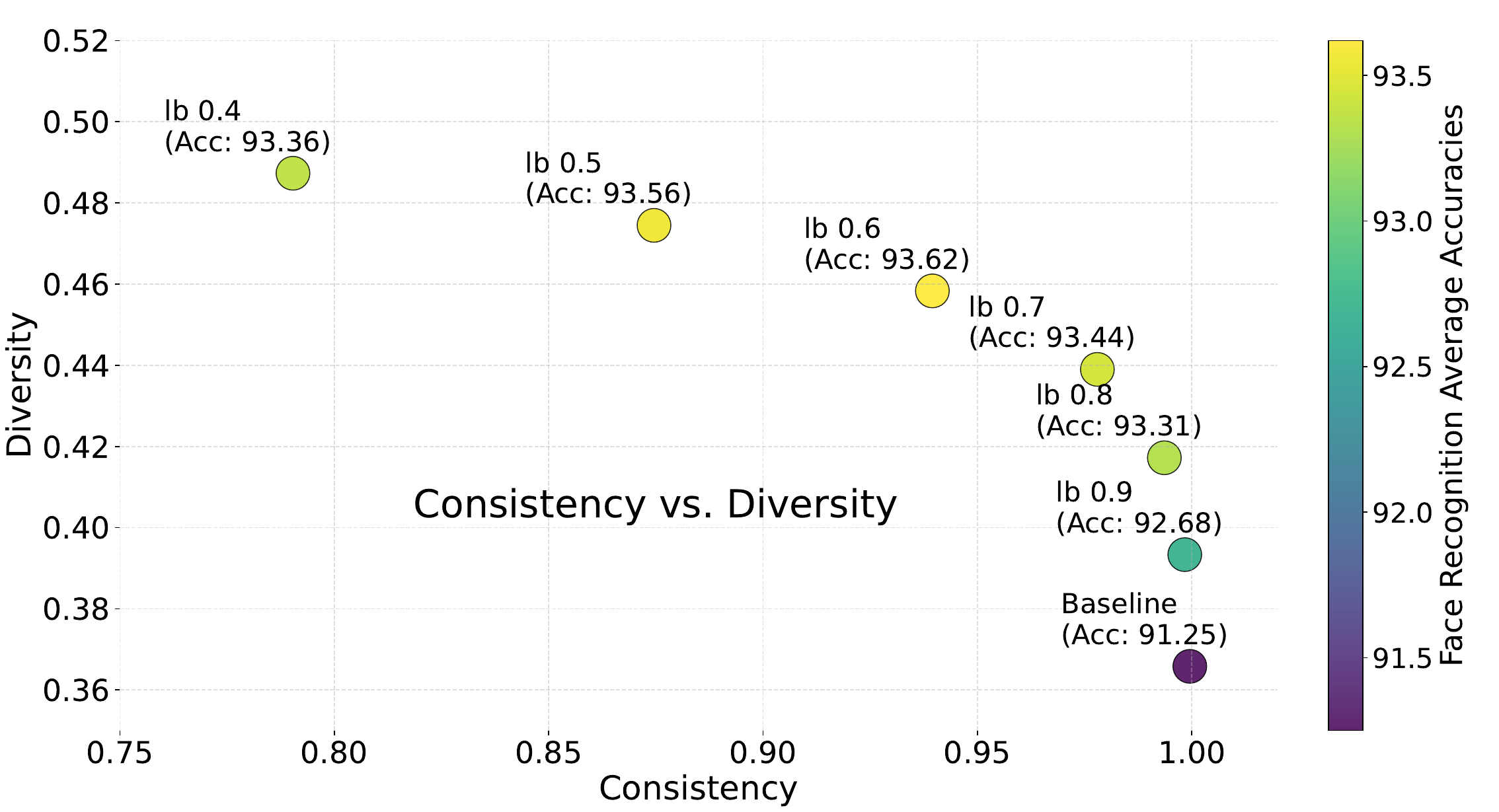} 
\vspace{-3mm}
\caption{FR average accuracies (as in Table \ref{tab:ablation_lb}) with respect to the intra-class consistency and diversity of baseline (C-WF) and \textsc{IDPerturb} of $lb \in [0.4,0.5,0.6,0.7,0.8,0.9]$.
}
\label{fig:tradeoff}
\vspace{-2mm}
\end{figure}

\begin{table}[ht!]
\centering
\caption{
Intra-class diversity and consistency on top of the Baseline(C-WF) reported as age entropy, facial expression (Exp.) entropy, head-pose STD, $D_{intra}$ and $C_{intra}$. Higher age and expression entropy, pose STD, and $D_{intra}$ indicate greater intra-class diversity. The first row shows authentic C-WF results. Baseline (C-WF) is generated from DM without \textsc{IDperturb}. The first part of the table shows results of different $lb$ and the second part presents the results of fixed $lb=0.6$ and different scale \( \omega \) of CFG.
}
\vspace{-1mm}
\resizebox{0.95\linewidth}{!}{%
\begin{tabular}{ccc|cccclcll}
\hline
\multicolumn{1}{l}{}       & \multicolumn{1}{l}{} & \multicolumn{1}{l|}{} & \multicolumn{5}{c}{Attributes}                                    & D-Intra & \multicolumn{1}{c}{C-Intra} & \multicolumn{1}{c}{FR-Perf. $\uparrow$} \\ \hline
Dataset                    & lb                    & scale                   & Age $\uparrow$ & \multicolumn{3}{c}{Pose $\uparrow$} & Exp. $\uparrow$ &         &                             &                              \\
                           &                       &                       &                & Yaw         & Pitch     & Roll      &            &         &                             &                              \\ \hline
C-WF                       & -                     & -                   & 0.354          & 23.479      & 9.069     & 5.154     & 0.589      & 0.423   & 0.948                       & 94.63                        \\ \hline
Baseline (C-WF)            & -                     & 1                     & 0.283          & 18.881      & 7.992     & 3.230     & 0.429      & 0.366   & 0.999                       & 91.25                        \\ \hline
\multirow{6}{*}{\textsc{IDperturb}} 
                           & 0.9                   & 1                     & 0.325          & 19.907      & 8.384     & 3.606     & 0.492      & 0.393   & 0.998                       & 92.68                        \\
                           & 0.8                   & 1                     & 0.369          & 20.899      & 8.727     & 4.029     & 0.538      & 0.417   & 0.994                       & 93.31                        \\
                           & 0.7                   & 1                     & 0.416          & 21.816      & 9.117     & 4.551     & 0.574      & 0.439   & 0.978                       & 93.44                        \\
                           & 0.6                   & 1                     & 0.461          & 22.735      & 9.467     & 5.124     & 0.603      & 0.458   & 0.939                       & 93.62                        \\
                           & 0.5                   & 1                     & 0.501          & 23.611      & 9.724     & 5.602     & 0.621      & 0.474   & 0.875                       & 93.56                        \\
                           & 0.4                   & 1                     & 0.538          & 24.279      & 9.957     & 6.040     & 0.636      & 0.487   & 0.790                       & 93.36                        \\ \hline
\multirow{6}{*}{\textsc{IDperturb}} 
        & 0.6 & 0 & 0.498 & 22.643 & 9.192 & 5.022 & 0.620 & 0.452 & 0.811 & 92.97 \\ 
        & 0.6 & 1 & 0.461 & 22.735 & 9.467 & 5.124 & 0.603 & 0.458 & 0.939 & 93.62 \\
        & 0.6 & 2 & 0.464 & 22.282 & 9.403 & 4.869 & 0.599 & 0.463 & 0.964 & 93.63 \\
        & 0.6 & 3 & 0.475 & 21.727 & 9.365 & 4.686 & 0.595 & 0.465 & 0.972 & 93.15 \\
        & 0.6 & 4 & 0.486 & 21.280 & 9.271 & 4.466 & 0.592 & 0.465 & 0.975 & 92.63 \\
        & 0.6 & 5 & 0.489 & 20.827 & 9.181 & 4.291 & 0.591 & 0.464 & 0.977 & 92.42 \\ \hline
\end{tabular}
}
\label{tab:intra_class_diversity}
\vspace{-4mm}
\end{table}

\section{Results}
\label{sec:result}
We first evaluate the impact of \textsc{IDperturb} on identity separability and intra-class diversity. We then assess the effectiveness of synthetic data generated using \textsc{IDperturb} as a FR training dataset across multiple FR benchmarks, reporting improvements compared to recent SOTA approaches.
\subsection{Impact of Identity Perturbation}
We begin by analyzing the effect of the $d$-dimensional cone boundary, defined by the lower bound hyperparameter \( \mathbf{lb} \), on identity-separability and downstream verification performance. Specifically, we generate datasets using different values of \( \mathbf{lb} \in \{0.9, 0.8, 0.7, 0.6, 0.5, 0.4\} \) on top of two pre-trained DMs: IDiff-Face trained on FFHQ and IDiff-Face trained on C-WF. These are compared against the baseline setting with no perturbation (\( \mathbf{lb} = 1.0\)). In these experiments, we fix the \( \omega = 1 \) of CFG, following \cite{DBLP:conf/iclr/LinHXMZD25}.

\textbf{Identity Separability Analysis}
We analyze the identity separability of the generated datasets under varying levels of $\mathbf{lb}$, as reported in the left part of Table~\ref{tab:ablation_lb}. The baseline (unperturbed) yields high identity separability, achieving an EER of 0.005, which is substantially lower than the 0.076 reported for authentic C-WF. This confirms that identity-conditioned DMs with CFG generate identity-consistent samples, supporting our statement in Methodology \ref{sec:methodology}.
However, this high separability also suggests that intra-class variation in the baseline synthetic data is limited, making the samples less challenging and potentially suboptimal for training robust FR models.
When applying \textsc{IDperturb}, we observe that decreasing \( \mathbf{lb} \) results in an expected increase in EER and a decrease in the mean of genuine scores. This trend reflects increased intra-class diversity, as the generated samples become more challenging. Importantly, even with strong perturbations, identity separability remains largely intact (e.g., \( \mathbf{lb} = 0.6 \) yielding EER = 0.082 vs. 0.076 reported for the authentic dataset C-WF), indicating that identity consistency is not severely compromised.

\begin{table*}[!ht]
\centering
\caption{
Identity separability of synthetic datasets (left side) and verification accuracy (\%)  of FR models trained on those datasets (right side).
We compare \textsc{IDperturb}-based synthetic datasets (generated using base DM trained on C-WF with varying scale $\omega \in [0, 5]$ values and fixed $d$-dimensional cone boundary $\mathbf{lb}=0.6$).}
\vspace{-2mm}
\resizebox{0.95\linewidth}{!}{%
\begin{tabular}{cc|ccccccccc||cccccc}
\hline
                                        &   & \multicolumn{8}{c}{\textbf{Identity-Separability of Training Data}}    & \phantom{abc}& \multicolumn{6}{c}{\textbf{FR Verification}}    \\
\hline
                                        &   & \multicolumn{2}{c}{\textbf{\multirow{2}{*}{Operation Metrics}}} & \phantom{abc} & \multicolumn{5}{c}{\textbf{Score Distributions}}    & \phantom{abc}& \multicolumn{6}{c}{\multirow{2}{*}{\textbf{Verification Benchmarks}}}    \\                              
  &&&&  \phantom{abc} &  \multicolumn{2}{c}{\textbf{genuine}} &   \multicolumn{2}{c}{\textbf{imposter}} &  \multicolumn{1}{c}{}&     &&&& \phantom{abc} &&      \\

\multicolumn{1}{c}{\textbf{Data Sampling Method}} & \textbf{Scale}& EER $\downarrow$  & FMR100 $\downarrow$ && G-mean & G-std & I-mean & I-std & FDR $\uparrow$ && LFW    $\uparrow$                   & AgeDB-30    $\uparrow$              & CFP-FP      $\uparrow$              & CALFW $\uparrow$                    & CP-LFW   $\uparrow$                 & Average     $\uparrow$                 \\ \hline
\multirow{6}{*}{\textsc{IDperturb}}               & 0 & 0.162 & 0.427 & \phantom{abc} & 0.185 & 0.124 & 0.006 & 0.058 & 1.718 & \phantom{abc}  & 99.25 & 92.58	& 92.31	& 93.53	& 87.17	& 92.97   \\
& 1  & 0.063 & 0.134 & \phantom{abc} &  0.338 & 0.162 & 0.008 & 0.059 & 3.664 & \phantom{abc}  & 99.40 & 93.20 & 93.61 &  93.50 & 88.37 & 93.62   \\
                                        & 2  & 0.043 & 0.079 & \phantom{abc} & 0.394 & 0.168 & 0.012 & 0.060 & 4.611 & \phantom{abc}  & 99.33 & 93.33 & 93.56 & 93.42 & 88.53 & 	\textbf{93.63}      \\
                                        & 3 & 0.036 & 0.064 & \phantom{abc} & 0.414 & 0.169 & 0.014 & 0.060 & 4.987 & \phantom{abc} & 99.22 & 92.83 & 92.66 & 93.20 & 87.83 & 93.15   \\
                                        & 4 & 0.034 & 0.060 & \phantom{abc} & 0.421 &	
                                        0.168 & 0.017 & 0.061 & 5.105 & \phantom{abc} & 99.07 & 91.90 & 92.04 & 92.70 & 87.42 & 92.63      \\
                                        & 5 & 0.034 & 0.059 & \phantom{abc} & 0.424 & 0.168 & 0.018 & 0.061 & 5.136 & \phantom{abc}  & 98.93 & 91.90 & 91.63 & 92.42 & 87.20 & 92.42      \\\hline
\end{tabular}}
\label{tab:ablation_scale}
\end{table*}

\textbf{Intra-Class Diversity and Consistency}
To further evaluate intra-class diversity and consistency, we report in Table \ref{tab:intra_class_diversity} the age entropy, facial expression entropy, STD of head-pose, intra-class perceptual diversity, and intra-class consistency. 
We observe that decreasing lb from 0.9 to 0.4 leads to a gradual increase in intra-class age and facial expression deviation (see also Figure \ref{fig:sample} C: baseline vs. lb 0.6), suggesting that angular perturbation successfully introduces naturalistic expression and age-related cues, without compromising identity coherence.  Similar observations can be made on head-pose, where the pose variation increases consistently with wider perturbation bounds, reaching up to STD of 23.611° yaw deviation at $lb = 0.5$. This indicates that \textsc{IDperturb} implicitly encourages pose diversification even though it operates solely in the identity-embedding space (see also Figure \ref{fig:sample} B and D: baseline vs. lb 0.5). As expected, the intra-class diversity ($D_{intra}$) increases monotonically with decreasing $lb$, confirming that \textsc{IDperturb} effectively expands the perceptual manifold of each identity Table \ref{tab:intra_class_diversity}.
A tradeoff between $D_{intra}$ and $C_{intra}$ is illustrated in Figure~\ref{fig:tradeoff} with FR performances of  models trained on \textsc{IDperturb} with the best performance achieved by \textsc{IDperturb} ($lb=0.6$), see Table \ref{tab:ablation_lb} for details evaluations.

\textbf{FR Verification Performance}
The right side of Table~\ref{tab:ablation_lb} presents the verification accuracy of FR models trained on datasets from \textsc{IDperturb}. Across all benchmarks, \textsc{IDperturb} consistently outperforms baseline models trained on data generated without identity perturbation.
For the baseline LDM model trained on  FFHQ, \textsc{IDperturb} achieves a peak average accuracy of 88.79\% at \( \mathbf{lb} = 0.5 \), compared to 86.58\% for the baseline (\( \mathbf{lb} = 1.0 \)). Notably, all tested values of \( \mathbf{lb} \in [0.9, 0.4] \) outperform the baseline.
Similarly, for the LDM model trained on  C-WF, \textsc{IDperturb} improves the average accuracy from 91.25\% to 93.62\%, with performance increasing as \( \mathbf{lb} \) decreases, until it peaks for \( \mathbf{lb} = 0.6 \). In all the following experiments
, we fix
Our method generalizes well across challenging benchmarks such as CALFW, AgeDB-30, and CP-LFW, which involve cross-pose and cross-age variations.

\begin{figure}[H]
\centering
\includegraphics[width=0.99\linewidth]{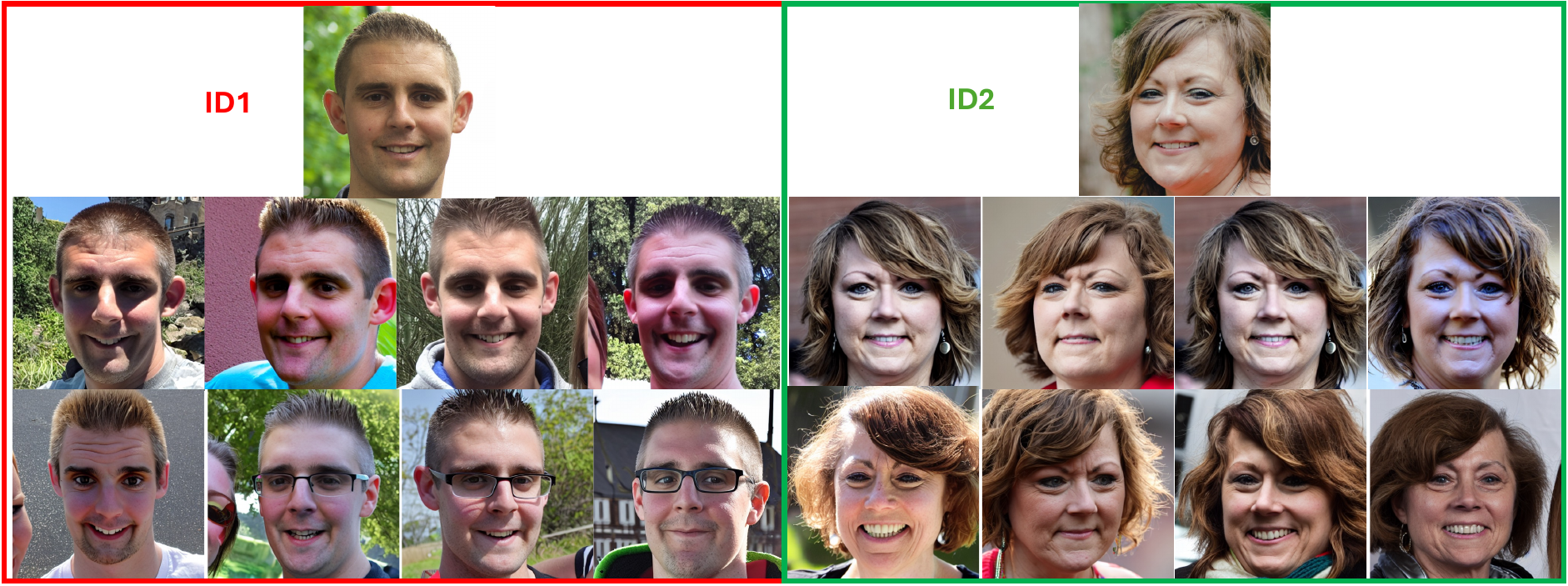} 
\vspace{-2mm}
\caption{Top: reference identity, first row is generated by Arc2Face and last row is generated by Arc2Face w/ \textsc{IDPerturb}. 
}
\label{fig:sames_arc}
\vspace{-3mm}
\end{figure}

\textbf{Generalizability of \textsc{IDperturb}} 
To demonstrate the generalizability of \textsc{IDperturb}, we adopt the official Arc2Face setup and pretrained model \cite{arc2face}. Since no synthetic data or identity embeddings are publicly available, we first train a PCA model following their protocol \cite{arc2face}. We then sample 10k novel identities and generate 50 images per identity, both with and without \textsc{IDperturb}. The FR evaluation results are reported in Table~\ref{tab:arc2face}, and Figure~\ref{fig:sames_arc} presents a visual comparison between Arc2Face and \textsc{IDperturb}.

\begin{table}[H]
\centering
\caption{Verification accuracy (\%) of FR models trained on data generated by  Arc2Face and Arc2Face w/ IDPERTURB.}
\label{tab:arc2face}
\vspace{-2mm}
\resizebox{0.95\linewidth}{!}{%
\begin{tabular}{lllllll}
\hline
                             & LFW   & AgeDB-30 & CFP-FP & CA-LFW & CP-LFW & Avg   \\ \hline
Arc2Face                     & 98.90 & 87.50 & 90.58  & 89.37  & 82.20  & 89.71 \\ \hline
Arc2Face w/ \textsc{IDPERTURB} & 98.93 & 89.81 & 92.57  & 90.90  & 84.40  & 91.32 \\
\hline
\end{tabular}}
\vspace{-2mm}
\end{table}

\subsection{Impact of CFG Strength}
To further understand the interaction between guidance strength and the model's adherence to the identity condition, we perform an ablation study over the CFG scale \( \omega \) in Equation~\ref{eq:cfg}. This study was performed on the C-FW dataset and, thus, we fixed the $d$-dimensional cone boundary based on the best performing setting, \( \mathbf{lb} = 0.6 \) (Table \ref{tab:ablation_lb}), while varying \( \omega \in \{0, 1, 2, 3, 4, 5\} \).
As shown in Table~\ref{tab:ablation_scale}, increasing \( \omega \) improves the model’s adherence to the identity condition \( \mathbf{v} \), as evidenced by reduced EER.  The increasing of \( \omega \) also   
enhances intra-class consistency while relatively reducing intra-class variation as illustrated in the second part of Table \ref{tab:intra_class_diversity}. However, stronger guidance also restricts sample diversity, resulting in reduced FR accuracy when training on highly guided synthetic datasets. The best verification performance is achieved at moderate guidance values \( \omega = 1 \) or \( \omega = 2 \).
In summary, these results highlight the importance of a trade-off between identity consistency, achieved via CFG, and intra-class variation, introduced through \textsc{IDperturb}. The combination of both is critical for generating effective training data for FR.

\begin{table*}[t!]
\centering
\caption{FR verification accuracies of \textsc{IDperturb} and SOTA.
We highlight the datasets used to train DGMs, FFHQ \cite{FFHQ} (70k images) vs. C-WF \cite{DBLP:journals/corr/YiLLL14a} (0.49M images) vs. WebFace4M  (WF4M) \cite{DBLP:conf/cvpr/ZhuHDY0CZYLD021} (4M images).
Under the same settings, i.e., DGM trained on large C-WF, \textsc{IDperturb} outperformed all competitors on the considered evaluation benchmarks. The results marked with a star (*) are reproduced in this work, as there was no publicly released FR model. Some of the previous works did not release their pre-trained model and their generated data and there were no reported results on the large-scale IJB-C \cite{DBLP:conf/icb/MazeADKMO0NACG18}. The first row presents the results achieved by authentic C-WF, which are provided as a reference. Best results for each DM setup are shown in \textbf{bold}, and second-best ones are \underline{underlined}.}
\vspace{-1mm}
\resizebox{0.95\textwidth}{!}{%
\begin{tabular}{c|c|cccccccc|cc}
\hline
  \multirow{2}{*}{\textbf{Method}} &
  \multirow{2}{*}{\textbf{Data Generation}} &
  \multirow{2}{*}{\textbf{DGMs Train Dataset}} &
  \multirow{2}{*}{\textbf{\#imgs (IDs × imgs/ID)}} &
  \multirow{2}{*}{\textbf{LFW}} &
  \multirow{2}{*}{\textbf{AgeDB}} &
  \multirow{2}{*}{\textbf{CFP-FP}} &
  \multirow{2}{*}{\textbf{CA-LFW}} &
  \multirow{2}{*}{\textbf{CP-LFW}} &
  \multirow{2}{*}{\textbf{Avg}} &
  \multicolumn{2}{c}{\textbf{IJB-C}} \\
  &&&&&&&&&&
  \textbf{$10^{-5}$} &
  \textbf{$10^{-4}$} \\ \hline
C-WF \cite{DBLP:journals/corr/YiLLL14a,DBLP:conf/iccv/BoutrosGKD23} &
  Authentic & 
  - &
  0.49M($\sim$10.5k × 47) & 
  99.55 &
  94.55 &
  95.31 &
  93.78 &
  89.95 &
  94.63 & 93.96 & 96.05
  \\ \hline
DigiFace \cite{DigiFace1M} WACV'22   & \multirow{3}{*}{Digital Rendering}                & - & 0.5M(10k × 50)   & 95.40 & 76.97 & 87.40 & 78.62 & 78.87 & 83.45 & - & - \\ 
DigiFace \cite{DigiFace1M}* WACV'22    &                & - & 0.5M(10k × 50)   & 91.15 & 74.00 & 82.93 & 75.30 & 73.40 & 79.36 & 30.01 &  44.78 \\ 
DigiFace \cite{DigiFace1M} WACV'22  &         & - & 1.2M(10k × 72 + 100k × 5)   &
96.17 &  81.10	&  89.81	& 82.55 & 82.23 &  86.37 &  -  &  - \\
\hline
IDnet \cite{DBLP:conf/cvpr/KolfREBKD23} CVPRW'23    &               \multirow{7}{*}{GAN-Based}                   & C-WF & 0.53M(10.5k × 50)  & 92.58 & 73.53 & 75.40  & 79.90 & 74.25 & 79.13 & 38.85 & 53.25 \\
SFace \cite{Boutros2022SFace}  IJCB'22           &                                  &  C-WF & 0.63M(10.5k × 60) & 91.87 & 71.68 & 73.86 & 77.93 & 73.20  & 77.71 & 12.70 & 19.87\\
SFace2+ \cite{10454585}   TBIOM'24                    &                                  & C-WF & 0.63M(10.5k × 60) & 95.60  & 77.37 &  77.11  & 83.40 & 74.60 & 81.62 & 0.85 & 5.36\\
SynFace \cite{DBLP:conf/iccv/QiuYG00T21} ICCV'21        &        & FFHQ &0.5M(10k × 50)   & 88.98 & -    & -     & -     & -     & -  & - & -   \\
SynFace (w/IM) \cite{DBLP:conf/iccv/QiuYG00T21} ICCV'21  &                                  & FFHQ &0.5M(10k × 50)   & 91.93 & 61.63 & 75.03 & 74.73 & 70.43 & 74.75 & - & - \\
USynthFace  \cite{DBLP:conf/fgr/BoutrosKFKD23} FG'23  &                                  & FFHQ & 0.4M(400k × 1)   & 92.23 & 71.62 & 78.56  & 77.05 & 72.03 & 78.30 & - & - \\
ExFaceGAN(Con)     \cite{ExFaceGAN}   IJCB'23       &                                  & FFHQ & 0.5M(10k × 50)   & 93.50  & 78.92  & 73.84 & 82.98  & 71.60 & 80.17 & 12.92 & 43.28 \\
 \hline

ID$^3$ \cite{DBLP:conf/nips/Xu0WXDJHM0DH24} NeurIPS'24 &   \multirow{17}{*}{Diffusion Model}                               & FFHQ & 0.5M(10k × 50)   & 97.28 & 83.78 & 85.00 & 89.30 & 77.13 & 86.50 & - & - \\
IDiff-Face  \cite{DBLP:conf/iccv/BoutrosGKD23}  ICCV'23                 & & FFHQ & 0.5M(10k × 50)   & \underline{98.00} & 86.43 & \textbf{85.47} & \underline{90.65} & 80.45 & \underline{88.20} & 20.60 & 62.60 \\ 


NegFaceDiff \cite{DBLP:journals/corr/abs-2508-09661} ICCV-W'25   && FFHQ & 0.5M(10k × 50)  & 97.60 &	\underline{86.53} &	\underline{85.33} &	90.28 & \underline{80.73} &	88.10 & \textbf{58.09} & \underline{73.93}\\

\textbf{\textsc{IDperturb} (Ours)}       &                                 & FFHQ & 0.5M(10k × 50)   & \textbf{98.55} & \textbf{88.85} & 84.27 & \textbf{91.42} & \textbf{80.85} & \textbf{88.79} & \underline{37.88} & \textbf{74.49} \\ \cline{1-1} \cline{3-12}
Arc2Face \cite{arc2face}  ECCV'24 &    & WF4M + FFHQ + CelebA & 0.5M(10k × 50) & 98.81 & 90.18 & 91.87 & 92.63 & 85.16 & 91.73 & - & - \\ 
HyperFace \cite{DBLP:conf/iclr/Otroshi-Shahreza25} ICLR'25 &                                  & WF4M + FFHQ + CelebA& 0.5M(10k × 50) & 98.50 & 86.53 & 88.83 & 89.40 & 84.23 & 89.29 & - & - \\
Vec2Face \cite{DBLP:conf/iclr/WuST0B25} ICLR'25 &                                  & WF4M & 0.5M(10k × 50) & 98.87 & \underline{93.12} & 88.97 & 93.57& 85.47 & 92.00 & - & - \\

DCFace  \cite{DBLP:conf/cvpr/Kim00023}  CVPR'23  &       & FFHQ + C-WF & 0.5M(10k × 50) & 98.55 & 89.70 & 85.33 & 91.60 & 82.62 & 89.56 & 60.80 & 74.63 \\ 
ID$^3$ \cite{DBLP:conf/nips/Xu0WXDJHM0DH24}  NeurIPS'24 &                                  & C-WF & 0.5M(10k × 50) & 97.68 & 91.00 & 86.84 & 90.73 & 82.77 & 89.80 & - & - \\

CemiFace \cite{DBLP:conf/nips/SunSPT24}  NeurIPS'24&                                  & C-WF & 0.5M(10k × 50) & 99.03 & 91.33 & 91.06 & \underline{92.42} & 87.65 & 92.30 & - & - \\

NegFaceDiff \cite{DBLP:journals/corr/abs-2508-09661} ICCV-W'25   && C-WF & 0.5M(10k × 50)  & 98.98 &	90.02 &	91.67 &	91.65	& \underline{88.82} &	92.23 & 77.38	& 86.11 \\

UIFace \cite{DBLP:conf/iclr/LinHXMZD25}  ICLR'25 &                                  & C-WF & 0.5M(10k × 50) & \underline{99.27} & 90.95 & \textbf{94.29} & 92.25 & \textbf{89.58} &\underline{ 93.27} & \underline{81.78}* & \underline{88.70}* \\

\textbf{\textsc{IDperturb} (Ours)}       &         & C-WF & 0.5M(10k × 50)  &
\textbf{99.40}&  	\textbf{93.20}	&  \underline{93.61}	& \textbf{ 93.50}&  	88.37&  	\textbf{93.62} &  \textbf{82.28} & \textbf{89.49} \\ 
\cline{1-1} \cline{3-12}

DCFace \cite{DBLP:conf/cvpr/Kim00023}  CVPR'23   &         & FFHQ + C-WF & 1.2M(20k × 50 + 40k × 5)   &
98.58 &  90.97	& 88.61 	&  92.82 & 85.07 &  91.21 &  -  &  - \\

CemiFace  \cite{DBLP:conf/nips/SunSPT24}  NeurIPS'24   &         & C-WF & 1.0M(20k × 50)  &
99.18 &  91.97	& 92.75 	& 93.01 & 88.42 & 93.07 &  -  &  - \\

Arc2Face \cite{arc2face}  ECCV'24    &         & WF4M + FFHQ + CelebA & 1.2M(20k × 50 + 40k × 5)  &
98.92 &  92.45	& 94.58 	& 93.33 &  86.45 &  93.14 &  -  &  - \\

Vec2Face \cite{DBLP:conf/iclr/WuST0B25}  ICLR'25   &         & WF4M & 1.0M(20k × 50 )   &
98.87 &  \underline{93.85}	& 89.87 	&  \underline{93.65} & 86.13 &  92.47 &  -  &  - \\

UIFace \cite{DBLP:conf/iclr/LinHXMZD25} ICLR'25  &         & C-WF & 1.0M(20k × 50)   &
\underline{99.22} &  92.45	&  \textbf{95.03}	& 93.18 & \textbf{90.42} &\underline{ 94.06} &  -  &  - \\

\textbf{\textsc{IDperturb} (Ours)}       &         & C-WF & 1.0M(20k × 50)  &

\textbf{99.48}	 & \textbf{94.03} & 	\underline{95.01} & 	\textbf{93.85} & 	\underline{90.01} & \textbf{94.48} &  \textbf{85.57} & \textbf{91.19}

\\ 
\hline
\hline
\end{tabular}%
}

\label{tab:sota}
\vspace{-2mm}
\end{table*}

\subsection{Comparison with State-of-the-Art}


We compare \textsc{IDperturb} with SOTA synthetic data generation methods, including digital rendering, GAN-based and DM-based approaches.
Compared to GAN or digital rendering (e.g., DigiFace-1M, IDNet, SFace, SFace2+), \textsc{IDperturb} achieves superior performance across all benchmarks, highlighting the advantages of DMs, especially when combined with geometry-aware identity sampling.
Among DM-based methods trained under comparable data regimes (FHHQ with 70K images) and by generating 0.5M images for FR training, \textsc{IDperturb} outperforms ID$^3$ and IDiff-Face on LFW, AgeDB, CALFW, and CP-LFW, and achieves nearly competitive performance on CFP-FP.
When the DM training data is scaled up (e.g.,  C-WF  with 0.49M or WebFace4M (WF4M) with 4M images), \textsc{IDperturb} remains highly competitive, ranking first on LFW, AgeDB, and CALFW, and second on CFP-FP and CP-LFW, slightly behind UIFace.
Overall, considering the average performance across five benchmarks, \textsc{IDperturb} achieves the highest average verification accuracy of 93.62\%, outperforming all compared synthetic-based FR systems under equivalent evaluation conditions.  Similar observation can be made on challenging IJB-C, where \textsc{IDperturb} outperforms other competitors.
In comparison to FR trained on authentic C-WF (average accuracy of 94.63\%), our synthetic \textsc{IDperturb} narrows the gap in constrained training settings with an average accuracy of 93.62\%.
We then generated 1.0M samples using \textsc{IDperturb} to train FR and compare to SOTA approaches that generated larger amounts of data to train FR systems. One can notice that \textsc{IDperturb} still ranks first with an average accuracy of 94.48\%, outperforming SOTA approaches. 

\section{Conclusion}
\vspace{-2mm}
We presented \textsc{IDperturb}, a geometry-driven sampling strategy designed for enhancing intra-class variation in identity-conditioned synthetic face generation. Unlike prior works that rely on auxiliary conditions, architectural modifications, or iterative sampling procedures, \textsc{IDperturb} perturbs identity embeddings within a constrained angular region of the hyper-sphere to produce diverse yet identity-consistent samples.
Through a series of experiments, we show that synthetic datasets generated with \textsc{IDperturb} yield strong performance across both small and large-scale face verification benchmarks, outperforming previous works on most of the considered benchmarks. These results validate the effectiveness of leveraging the geometric structure of identity embeddings to induce diversity in the generated samples, improving the generalizability of FR models without sacrificing semantic coherence.


\section*{Acknowledgment}
This research work has been funded by the German Federal Ministry of Education and Research and the Hessen State Ministry for Higher Education, Research and the Arts within their joint support of the National Research Center for Applied Cybersecurity ATHENE.

{
    \small
    \bibliographystyle{ieeenat_fullname}
    \bibliography{main}
}


\end{document}


\maketitle

\section{Supplementary Material}
\begin{figure*}
    \centering
    \includegraphics[width=0.75\linewidth]{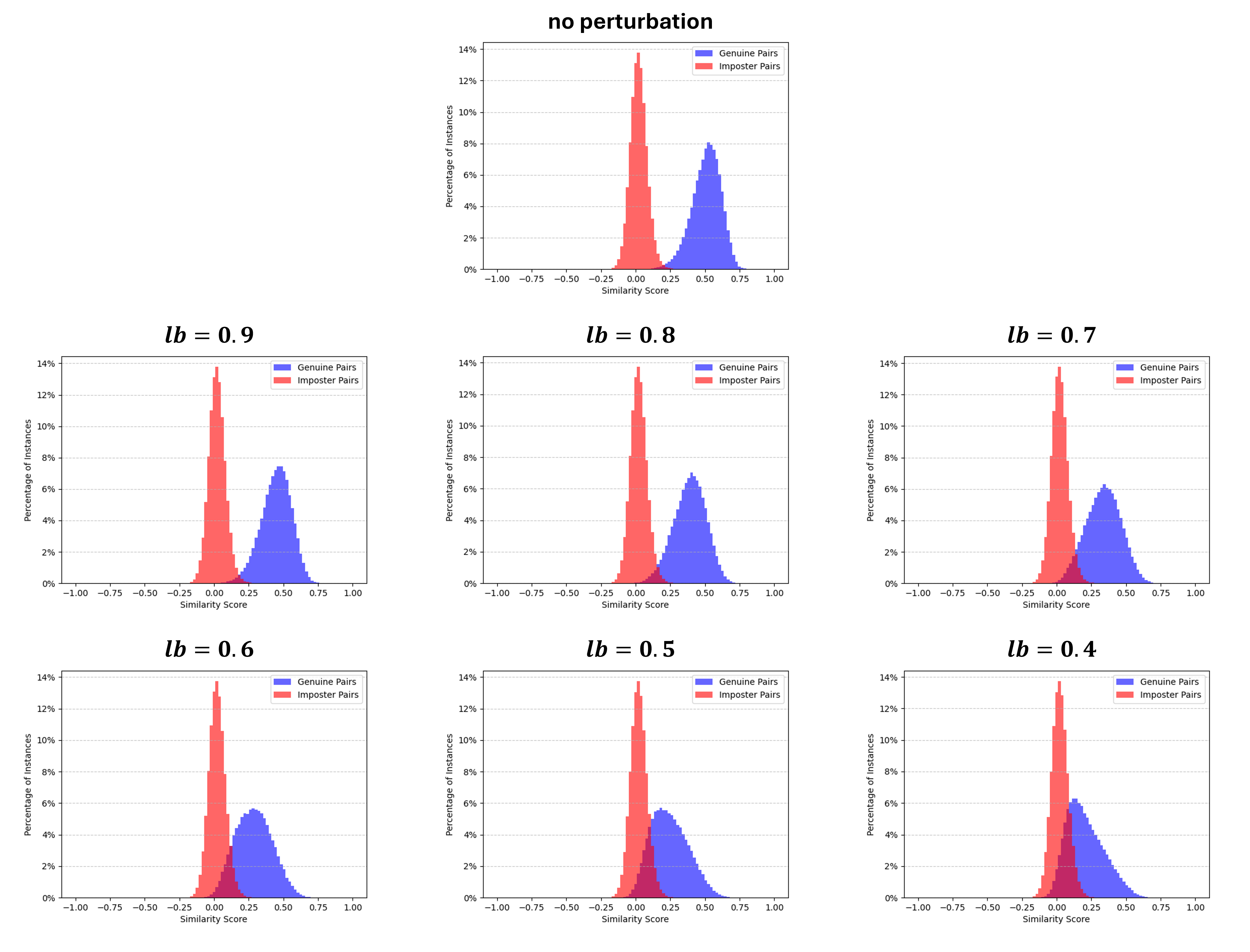}
    \caption{Histograms of the genuine and impostor score distributions for IDiff-Face \cite{DBLP:conf/iccv/BoutrosGKD23} (baseline, no perturbation), and the proposed \textsc{IDperturb} method trained on the FFHQ dataset. The \textsc{IDperturb} results correspond to data generated by varying the $d$-dimensional cone boundary $\mathbf{lb} \in [0.4,0.9]$. One can notice a decrease in identity-separability when the value of data $\mathbf{lb}$ decreases.These results correspond to a visual representation of the values presented in the first section of Table 1 of the main document.}
    \label{fig:id_sep_ffhq}
\end{figure*}

\begin{figure*}
    \centering
    \includegraphics[width=0.75\linewidth]{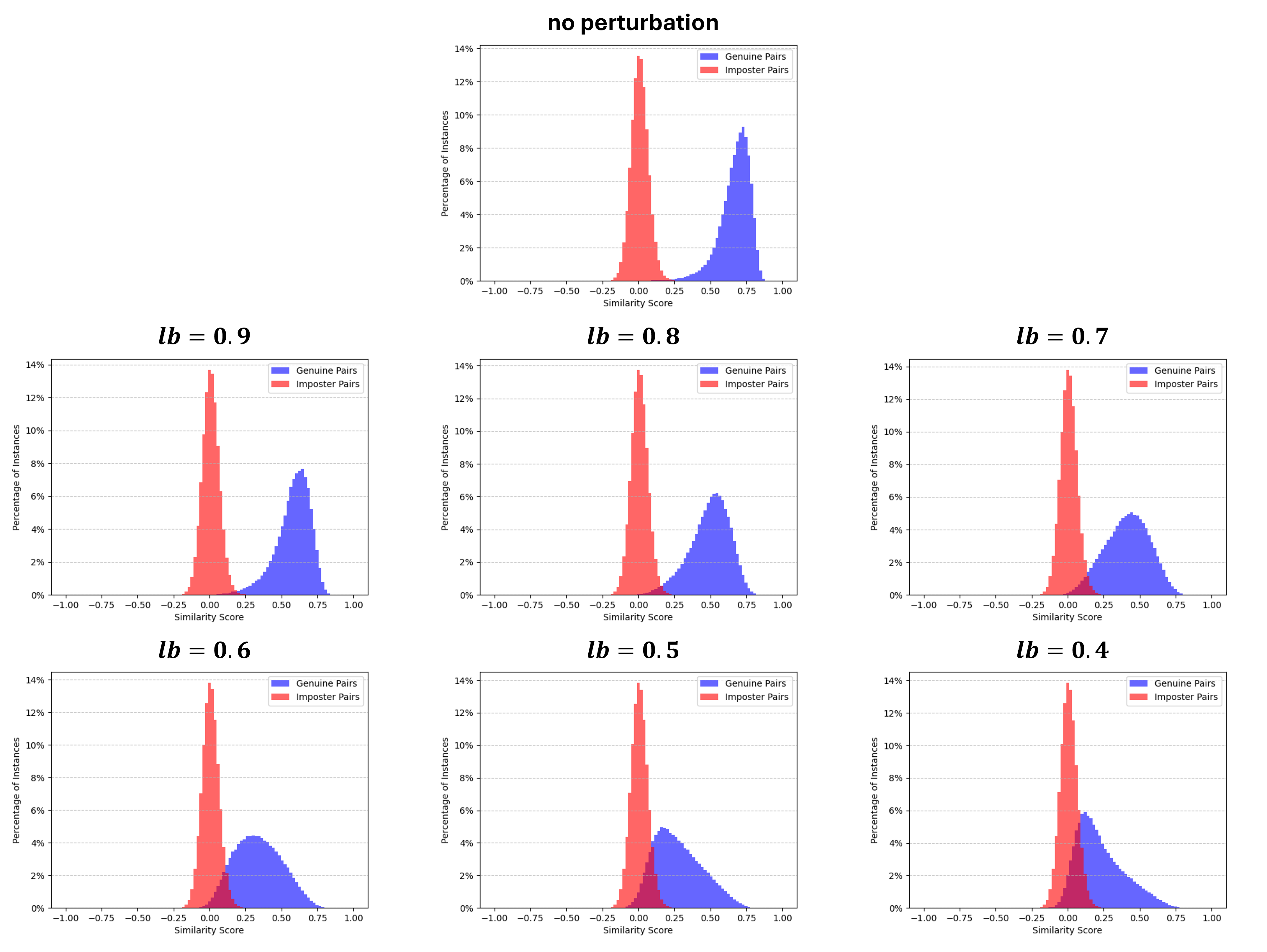}
    \caption{Histograms of the genuine and impostor score distributions for IDiff-Face \cite{DBLP:conf/iccv/BoutrosGKD23} (baseline, no perturbation), and the proposed \textsc{IDperturb} method trained on the CASIA dataset. The \textsc{IDperturb} results correspond to data generated by varying the $d$-dimensional cone boundary $\mathbf{lb} \in [0.4,0.9]$. One can notice a decrease in identity-separability when the value of data $\mathbf{lb}$ decreases.These results correspond to a visual representation of the values presented in the second section of Table 1 of the main document.}
    \label{fig:id_sep_CASIA_lb}
\end{figure*}

\begin{figure*}
    \centering
    \includegraphics[width=0.75\linewidth]{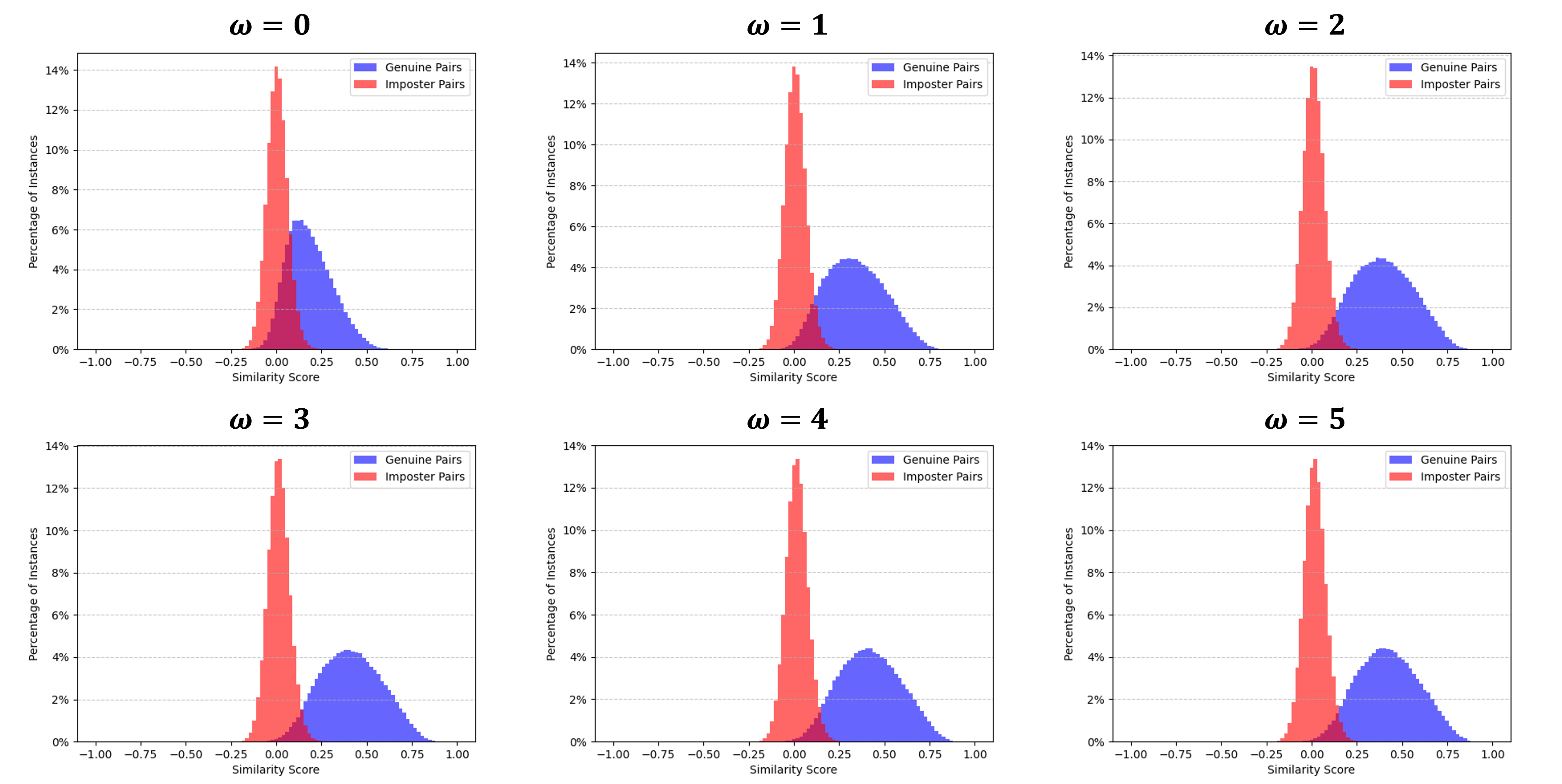}
    \caption{Histograms of the genuine and impostor score distributions for the proposed \textsc{IDperturb} method trained on the CASIA dataset. The \textsc{IDperturb} results correspond to data generated by varying scale $\omega \in [0, 5]$ and fixing the $d$-dimentional cone boundary as $\mathbf{lb}=0.6$. One can notice a increase in identity-separability when the value of data $\omega$ increases.These results correspond to a visual representation of the values presented in Table 2 of the main document.}
    \label{fig:id_sep_CASIA_w}
\end{figure*}

\begin{figure*}
    \centering
    \includegraphics[width=0.85\linewidth]{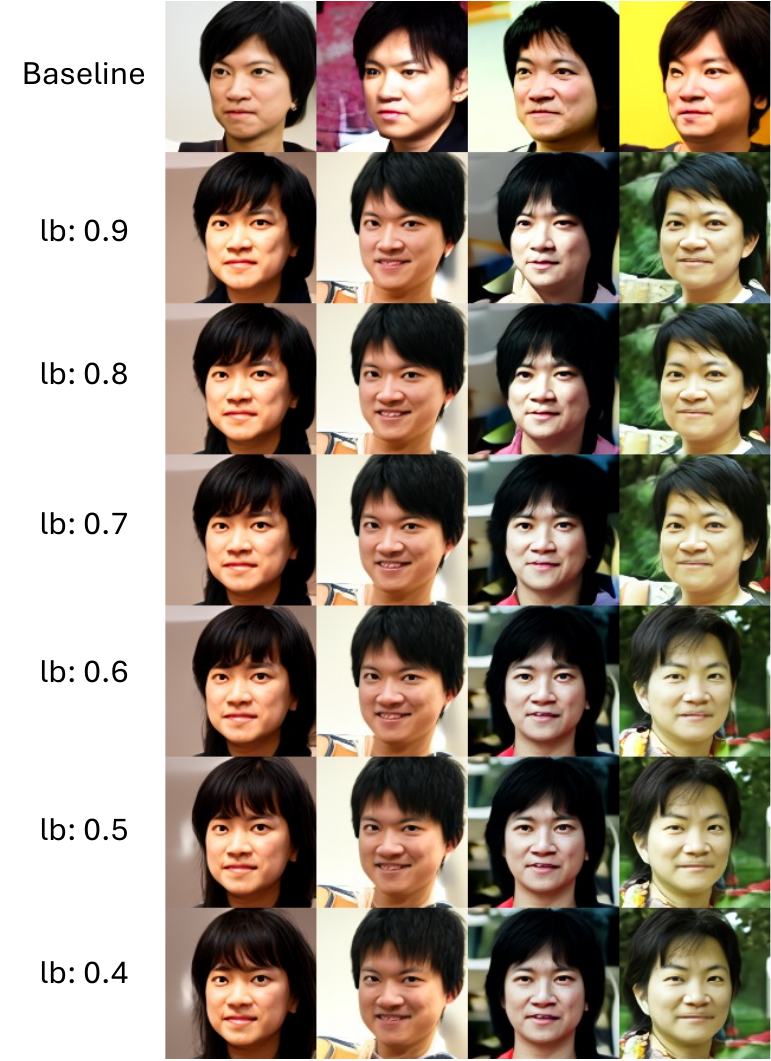}
    \caption{Samples generated from baseline model (without perturbation) and the ones generated using \textsc{IDperturb} with $lb \in [0.4,0.9]$ }
    \label{fig:samples001}
\end{figure*}

\begin{figure*}
    \centering
    \includegraphics[width=0.85\linewidth]{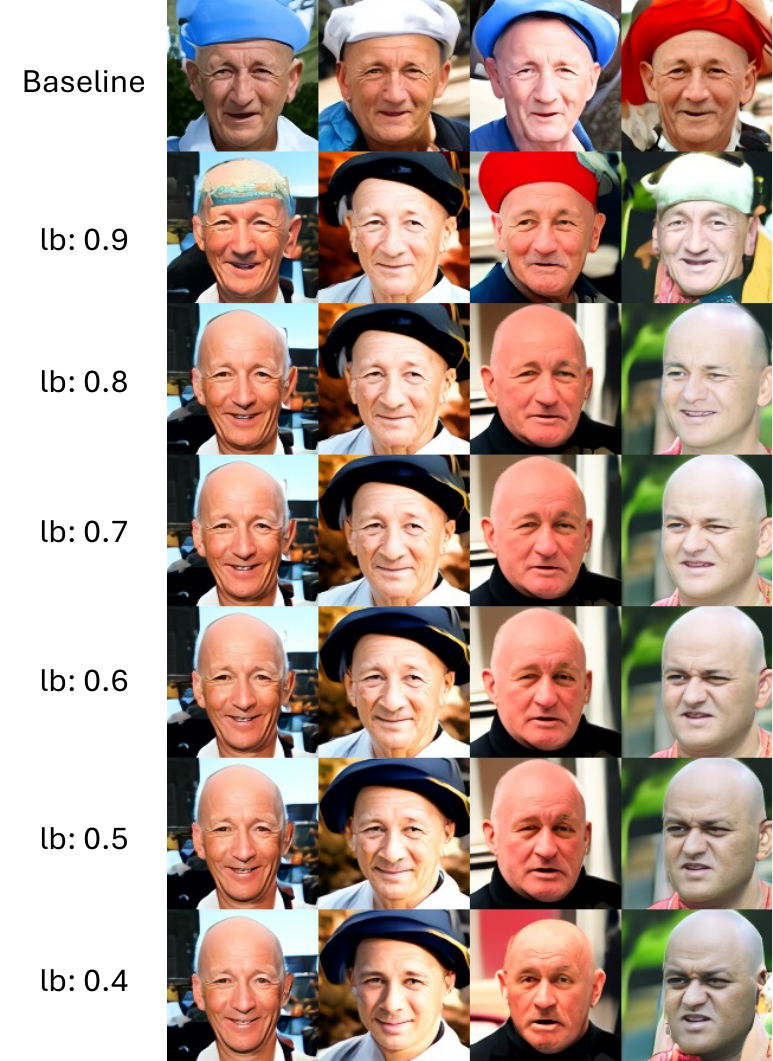}
    \caption{Samples generated from baseline model (without perturbation) and the ones generated using \textsc{IDperturb} with $lb \in [0.4,0.9]$ }
    \label{fig:samples002}
\end{figure*}

\begin{figure*}
    \centering
    \includegraphics[width=0.85\linewidth]{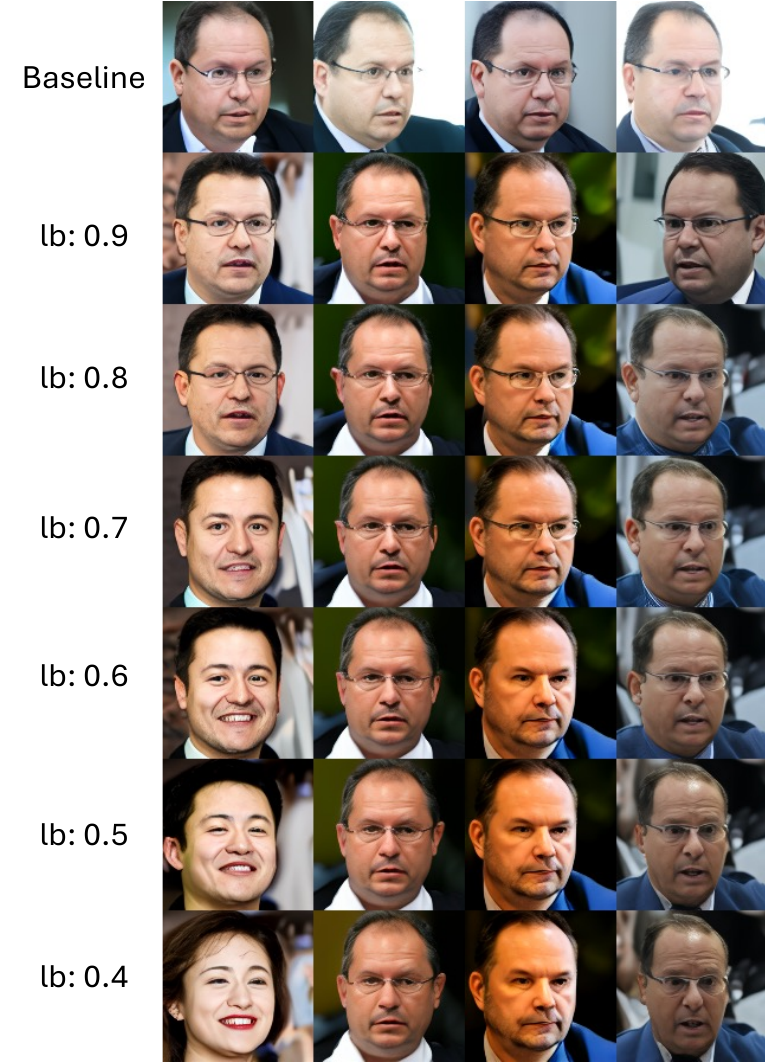}
    \caption{Samples generated from baseline model (without perturbation) and the ones generated using \textsc{IDperturb} with $lb \in [0.4,0.9]$ }
    \label{fig:samples003}
\end{figure*}

\begin{figure*}
    \centering
    \includegraphics[width=0.85\linewidth]{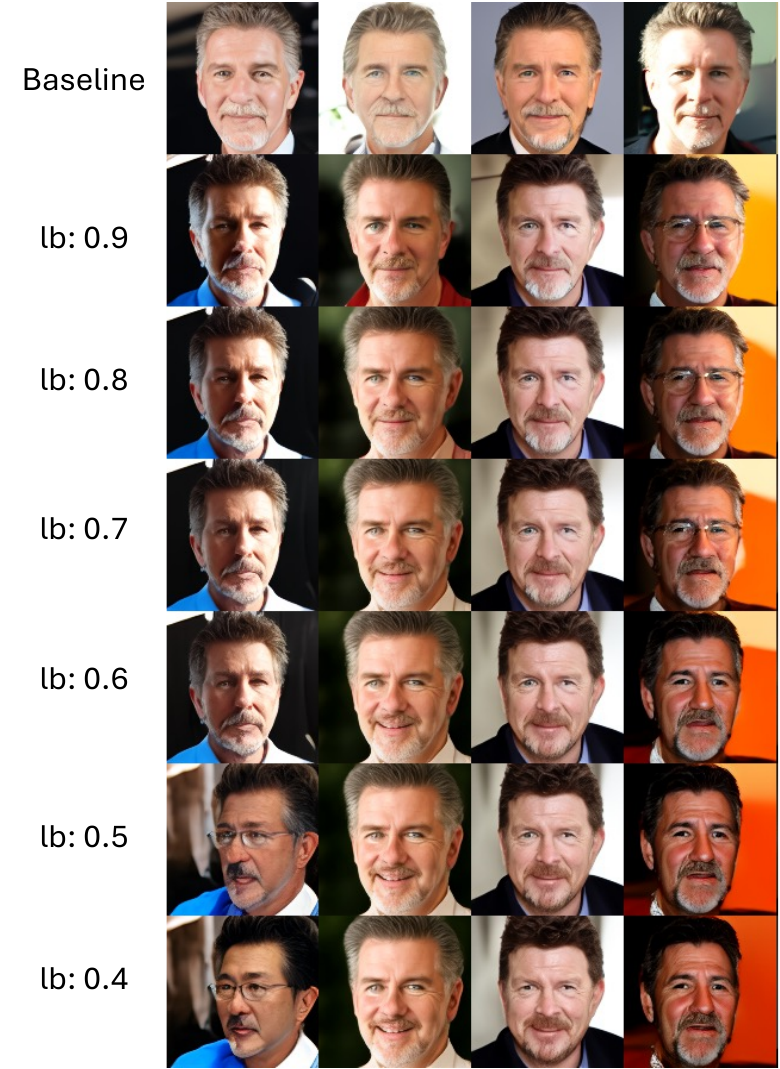}
    \caption{Samples generated from baseline model (without perturbation) and the ones generated using \textsc{IDperturb} with $lb \in [0.4,0.9]$ }
    \label{fig:samples004}
\end{figure*}

This supplementary material complements the main submission by providing:
\begin{enumerate}
  \item Appendix A (Sec. \ref{sec:app_a}, Table \ref{tab:noise}): Theoretical justification of \textsc{IDperturb}  
  
  \item Appendix B (Sec. \ref{sec:app_b}, Table \ref{tab:rfw}): Bias assessment in synthetic-based face recognition (FR) 
  
  \item Appendix C  (Sec. \ref{sec:app_c}, Table \ref{tab:loss_func}): Face recognition performance using different loss functions and network architectures
  
  \item  Appendix D (Sec \ref{sec:app_d}, Table \ref{tab:diversity_consistency_ffhq}): Intra-class diversity and consistency on top of the Baseline(FFHQ) reported as age entropy, facial expression (Exp.) entropy, head-pose STD, $D_{intra}$ and $C_{intra}$. These results are complementary to those in Table 2 (main submission). The results are reported on datasets generated with $lb \in [0.9,0.8,0.7,0.6,0.5,0.4]$.

  
  \item Appendix E (Sec. \ref{sec:app_e}, Table \ref{tab:diversity_consistency_sota}): Intra-class diversity and consistency of our \textsc{IDperturb} compared to several State-of-the-art (SOTA) approaches 
  
  \item Appendix F (Sec. \ref{sec:app_f}, Figure \ref{fig:id_sep_ffhq}, Figure \ref{fig:id_sep_CASIA_lb} and Figure \ref{fig:id_sep_CASIA_w}): Histogram of Genuine-Imposter score distributions are shown in Figure \ref{fig:id_sep_ffhq}, Figure \ref{fig:id_sep_CASIA_lb} and Figure \ref{fig:id_sep_CASIA_w}. These plots correspond to the results presented in the main submission in Tables 1 (Figure \ref{fig:id_sep_ffhq}, Figure \ref{fig:id_sep_CASIA_lb} ) and 3 Figure \ref{fig:id_sep_CASIA_w}, respectively.
  
  \item Appendix G (Sec. \ref{sec:app_g}): Use of existing assets and face recognition evaluation benchmarks.
  \item  Appendix H (Sec. \ref{sec:app_h}): Social Impact 
  \item Samples of the generated data in Figure \ref{fig:samples001}, \ref{fig:samples002}, \ref{fig:samples003} and \ref{fig:samples004}. These samples  complement the one provided in Figure 2 of the main submission. 
  
  \item  The code to reproduce the results of the main submission is provided as a zip file CVPR2026\_code.zip

\end{enumerate}

\section{Appendix A: Theoretical justification of \textsc{IDperturb}}
\label{sec:app_a}
\paragraph{Theoretical Justification for Angular Sampling}

Let \( \mathbf{v} \in  \mathbb{R}^d \) be a normalized identity embedding, i.e., \( \|\mathbf{v}\|_2 = 1 \). Modern face recognition (FR) models rely on angular distance between such embeddings to measure identity similarity. Given two embeddings \( \mathbf{v} \) and \( \mathbf{\tilde{v}} \), their cosine similarity is given by:

\begin{equation}
    \cos(\theta_{v,\tilde{v}}) = \mathbf{v}^\top \mathbf{\tilde{v}}.
\end{equation}

We define the angular neighborhood of \( \mathbf{v} \) as the spherical cap:
\begin{equation}
    \mathcal{C}_{lb}(\mathbf{v}) = \left\{ \mathbf{\tilde{v}} \in \mathbb{R}^d \;\middle|\; \mathbf{v}^\top \mathbf{\tilde{v}} \geq lb \right\},
\end{equation}
where \( lb \in [0, 1] \) is a lower bound on cosine similarity (i.e., an upper bound on angular deviation \( \theta_{\max} = \arccos(lb) \)). Sampling within \( \mathcal{C}_{lb}(\mathbf{v}) \) ensures that perturbed embeddings remain semantically close to the original identity.

This geometric constraint:
\begin{itemize}
    \item Maintains the unit norm of embeddings, preserving their location on the identity hypersphere.
    \item Ensures consistent and interpretable control over identity similarity, as cosine similarity correlates with FR decision boundaries.
\end{itemize}

This theoretical alignment explains the empirical success of angular perturbation, as shown in the main submission. It leverages the intrinsic structure of FR models' embedding space to inject meaningful diversity without significantly degrading identity coherence.

\paragraph{Comparison with Random Noise Perturbation.}
An alternative approach would be to add random noise in the Euclidean space:
\begin{equation}
\tilde{\mathbf{v}} = \mathbf{v} + \epsilon, \quad \epsilon \sim \mathcal{N}(0,  \mathbf{I}),
\end{equation}
followed by normalization. However, this method introduces limitations such as inconsistent angular similarity, as Euclidean perturbations do not provide a guaranteed bound on angular deviation, leading to unpredictable identity similarity. This is supported by the results presented in Table \ref{tab:noise}, which reveal the poor FR recognition performance achieved when using datasets generated by models trained with the addition of random noise in comparison with IDiff-Face (baseline) and our \textsc{IDperturb}. It can also be observed that adding noise as context perturbation resulted in reduced identity-separability (high EER), which is coherent with the absence of a bounded context variation associated with this method. 

\begin{table*}[ht!]
\centering
\caption{
\textsc{IDperturb} vs. Noise perturbations. 
We compare \textsc{IDperturb} with a naive baseline in which identity embeddings are perturbed by adding Gaussian noise directly, without geometric constraints. For each of the base diffusion models trained on FFHQ and Casia-WebFace, we report results for three settings: (i) the unperturbed baseline, (ii) our proposed \textsc{IDperturb}, and (iii) random noise perturbations. We evaluate both the identity separability of the generated datasets and the face verification accuracy (\%) of FR models trained on them. The results clearly show that while \textsc{IDperturb} introduces diversity without sacrificing identity semantics, random noise perturbations severely degrade identity consistency, as evidenced by the poor separability metrics and significantly lower verification performance. These findings highlight the importance of controlled, geometry-aware perturbations for generating effective synthetic training data in face recognition. These empirical results support our second justification in Appendix A.
}
\resizebox{\linewidth}{!}{%
\begin{tabular}{c|ccccccc|cccccc}
\hline
                              & \multicolumn{7}{c|}{\textbf{Identity-Separability of Training Data}}                                                                                            & \multicolumn{6}{c}{\textbf{FR Verification}}                          \\ \hline
                              & \multicolumn{2}{c}{\multirow{2}{*}{\textbf{Operation Metrics}}} & \multicolumn{5}{c|}{\textbf{Score Distributions}}                                             & \multicolumn{6}{c}{\multirow{2}{*}{\textbf{Verification Benchmarks}}} \\
                              & \multicolumn{2}{c}{}                                            & \multicolumn{2}{c}{\textbf{genuine}} & \multicolumn{2}{c}{\textbf{imposter}} &                & \multicolumn{6}{c}{}                                                  \\
\textbf{Data Sampling Method} & EER $\downarrow$              & FMR100 $\downarrow$             & mean              & std              & mean               & std              & FDR $\uparrow$ & LFW    & AgeDB-30  & CFP-FP  & CALFW  & CP-LFW  & Average $\uparrow$  \\ \hline
Baseline (FFHQ)               & 0.005                         & 0.004                           & 0.509             & 0.104            & 0.023              & 0.081            & 13.680         & 97.60  & 84.10     & 83.36   & 89.08  & 78.77   & 86.58               \\ \hline
\textsc{IDperturb}              & 0.124                         & 0.332                           & 0.245             & 0.133            & 0.023              & 0.060            & 2.316          & 98.55  & 88.85     & 84.27   & 91.42  & 80.85   & \textbf{88.79}      \\ \hline
Noise                         & 0.498                         & 0.949                           & 0.009             & 0.060            & 0.008              & 0.060            & 0              & 57.48  & 50.97     & 56.76   & 53.07  & 51.45   & 53.94               \\ \hline \hline
Baseline (C-WF)               & 0.003                         & 0.001                           & 0.670             & 0.107            & 0.010              & 0.060            & 29.116         & 98.75  & 88.85     & 91.61   & 90.90  & 86.15   & 91.25               \\ \hline
\textsc{IDperturb}               & 0.063                         & 0.134                           & 0.338             & 0.162            & 0.008              & 0.059            & 3.664          & 99.40  & 93.20     & 93.61   & 93.50  & 88.37   & \textbf{93.62}      \\ \hline
Noise                         & 0.498                         & 0.990                           & 0.009             & 0.060            & 0.008              & 0.060            & 0              & 64.58  & 56.35     & 63.56   & 54.40  & 54.55   & 58.69               \\ \hline
\end{tabular}}
\label{tab:noise}
\end{table*}

\paragraph{Theoretical Properties of Angular Sampling}


\begin{lemma}[Norm Preservation]
Let \( \tilde{\mathbf{v}} = \cos(\theta) \mathbf{v} + \sin(\theta) \mathbf{u} \), where \( \|\mathbf{v}\| = \|\mathbf{u}\| = 1 \), and \( \langle \mathbf{v}, \mathbf{u} \rangle = 0 \). Then \( \tilde{\mathbf{v}} \) satisfies:
\[
\| \tilde{\mathbf{v}} \| = 1
\]
\end{lemma}

\begin{proof}
\begin{align*}
\| \tilde{\mathbf{v}} \|^2 &= \| \cos(\theta)\mathbf{v} + \sin(\theta)\mathbf{u} \|^2 \\
&= \cos^2(\theta)\|\mathbf{v}\|^2 + \sin^2(\theta)\|\mathbf{u}\|^2 + 2\cos(\theta)\sin(\theta)\langle \mathbf{v}, \mathbf{u} \rangle \\
&= \cos^2(\theta) + \sin^2(\theta) + 0 = 1
\end{align*}
\end{proof}

\begin{lemma}[Cosine Similarity Control]
Given \( \tilde{\mathbf{v}} = \cos(\theta) \mathbf{v} + \sin(\theta) \mathbf{u} \) with \( \mathbf{u} \perp \mathbf{v} \), it holds that:
\[
\langle \tilde{\mathbf{v}}, \mathbf{v} \rangle = \cos(\theta)
\]
\end{lemma}

\begin{proof}
\begin{align*}
\langle \tilde{\mathbf{v}}, \mathbf{v} \rangle &= \langle \cos(\theta)\mathbf{v} + \sin(\theta)\mathbf{u}, \mathbf{v} \rangle \\
&= \cos(\theta)\langle \mathbf{v}, \mathbf{v} \rangle + \sin(\theta)\langle \mathbf{u}, \mathbf{v} \rangle = \cos(\theta)
\end{align*}
\end{proof}

\noindent \textbf{Avoiding Identity Overlap}

\textbf{Theorem 1}
Let \( \mathcal{I} = \{ \mathbf{v}_1, \ldots, \mathbf{v}_k \} \subset \mathbb{R}^d \) be a set of unit-normalized identity vectors. To ensure that a perturbed vector \( \tilde{\mathbf{v}} \) sampled from \( \mathcal{C}_{lb}(\mathbf{v}_i)\) remains closer to \( \mathbf{v}_i \) than any \( \mathbf{v}_l, l \ne i \), it suffices that:

\begin{equation}
   \mathbf{lb} \leftarrow \max\left(\mathbf{lb}, \max_{l \ne i} \text{cos} \left(\frac{\angle(\mathbf{v_i}, \mathbf{v_l})}{2} \right)\right).
   \label{eq:lb_cond}
\end{equation}



\begin{proof}
Let $v_j$ be the closest identity context to $v_i$ in $\mathbb{R}^d$, that is $\angle(v_i,v_j) \leq \angle(v_i, v_l), v_l \in \mathcal{I} \land l \neq i$. Equation \ref{eq:lb_cond} can be rewritten as:

\begin{align*}
\mathbf{lb} \geq \text{cos} \left(\frac{\angle(\mathbf{v_i}, \mathbf{v_j})}{2} \right) \leftrightarrow \theta_{\max} \leq \angle\frac{(\mathbf{v_i}, \mathbf{v_j})}{2} .
   \label{eq:lb_cond}
\end{align*}

Considering that $\angle(v_i, \hat{v}) \leq \theta_{\max}$, it can be further derived that

\begin{align*}
   \angle(v_i, v_j) \geq 2 \times \angle(v_i, \hat{v})
\end{align*}

Taking the triangle inequality for angular distances into account:

\begin{align*}
    \angle (v_i, v_j) \leq \angle(v_i, \hat{v}) + \angle(\hat{v}, v_j) \leftrightarrow \\
    2 \times \angle(v_i, \hat{v})  \leq \angle(v_i, \hat{v}) + \angle(\hat{v}, v_j) \leftrightarrow \\
    \angle(v_i, \hat{v})  \leq \angle(v_j, \hat{v}),    
\end{align*}
which concludes the proof. Given the validity of the theorem on the most extreme scenario where $\angle(v_i,v_j) \leq \angle(v_i, v_l), v_l \in \mathcal{I} \land l \neq i$, it is trivial to derive its validity for all $l \neq i$.


\end{proof}
\section{Appendix B: Racial Bias in synthetic-based FR}
\label{sec:app_b}
Our FR models trained on the \textsc{IDperturb} dataset were evaluated on the Racial Faces in the Wild (RFW) dataset \cite{RFW} to assess model bias and performance across different demographic groups.  The RFW dataset contains four testing subsets corresponding to African, Asian, Caucasian, and Indian groups and includes approximately 80k images from 12k identities. We report the results as verification accuracies in (\%) on each subset and as average accuracies to evaluate general recognition performance on the considered evaluation benchmarks. To evaluate the bias, we report the STD between all subsets and the SER, which is given by $\frac{max_gError_g}{min_gError_g}$, where $g$ represents the demographic group, as reported in \cite{PAMI_BIAS, DBLP:journals/corr/abs-1911-10692}. A higher STD value indicates more bias across demographic groups and vice versa. For SER, models that achieved values closer to 1 are less biased. From the reported results in Table \ref{tab:rfw}, comparing the bias assessment of FR models trained on the  \textsc{IDperturb} dataset and SOTA data generation methods, we observe that  \textsc{IDperturb} achieves lower STD and SER compared to the other considered base diffusion models trained on Casia-WebFace, while maintaining comparable STD and SER to FFHQ-based methods. Additionally, \textsc{IDperturb} achieves notably higher average accuracy on RFW compared to the other considered approaches.

\begin{table*}[ht!]
\centering
\caption{Bias assessment of FR models trained on  \textsc{IDperturb} dataset and SOTA data generation methods. The ethnicity-specific results correspond to verification accuracies in \% on each ethnicity subset of RFW \cite{RFW}. STD and SER are the bias metrics that can be calculated based on these verification accuracies. 
The best results for each metric are highlighted in bold. The results in the first row are obtained by training FR on authentic CASIA-WebFace, followed by FR model trained on digital rendering (DigiFace). We then present the results of the FR models trained on synthetically generated data, where the base generative model is trained on FFHQ, followed by a base generative model trained on CASIA-WebFace. }
\begin{tabular}{ccccc|c||cc}
\hline
Models    & Indian & Caucasian & Asian & African & Avg. & STD & SER \\ \hline
CASIA-WebFace \cite{DBLP:journals/corr/YiLLL14a} & 88.50 & 93.38 & 85.38 & 86.62 & 88.47 & 3.52 & 2.21 \\ \hline
DigiFace1M \cite{DigiFace1M}   & 70.33  &  72.18    & 69.97 & 65.55   &  69.51  &  2.81   &  1.23 \\ \hline

ExFaceGAN (FFHQ) \cite{ExFaceGAN} & 71.58  & 73.80     & 71.07 & 62.20   &  69.66  &  5.11   & 1.44  \\ 
IDiff-Face (FFHQ) \cite{DBLP:conf/iccv/BoutrosGKD23} & 81.50  & 85.37     & 79.77 & 75.78   &   80.61  &   3.98  &  1.66  \\ 
IDperturb (FFHQ) (Ours) & 84.25 & 86.78 & 81.18 & 77.17 & 82.35	& 4.14 & 1.73 \\ \hline

SFace (Casia-WebFace) \cite{Boutros2022SFace}   & 67.07  & 73.15     & 68.07 & 63.57   &  67.97  &   3.96  &   \textbf{1.36}  \\

IDNet (Casia-WebFace) \cite{DBLP:conf/cvpr/KolfREBKD23}  & 71.93  & 76.17     & 70.85 & 64.07   &  70.76  &  5.01   & 1.51  \\


DCFace (Casia-WebFace) \cite{DBLP:conf/cvpr/Kim00023}   & 81.63  & 86.23     & 79.05 & 74.05   &  80.24 &  5.08   &  1.88  \\
UIFace (Casia-WebFace) \cite{DBLP:conf/iclr/LinHXMZD25}  & 88.58  & \textbf{91.68} & 85.32 & 83.57 & 87.29 & 3.59 & 1.97 \\ 

IDperturb (Casia-WebFace) (Ours) & \textbf{88.83} & 91.60 & \textbf{86.23} & \textbf{85.80} & \textbf{88.12} & \textbf{2.68} & 1.69 \\ \hline 

\end{tabular}
\label{tab:rfw}

\end{table*}

\begin{table*}[ht!]
\centering
\caption{
Intra-class diversity and consistency on top of the Baseline(FFHQ) reported as age entropy, facial expression (Exp.) entropy, head-pose STD, $D_{intra}$ and $C_{intra}$. Higher age and expression entropy, and pose STD, and  $D_{intra}$ indicate greater intra-class diversity. Baseline (FFHQ) is generated from the DM without \textsc{IDperturb}. These results are complementary to those reported in the main submission Table 2.
}
\begin{tabular}{cc|cccclcll}
\hline
\multicolumn{1}{l}{}                & \multicolumn{1}{l|}{} & \multicolumn{5}{c}{Attributes}                               & D-Intra & \multicolumn{1}{c}{C-Intra} & \multicolumn{1}{c}{FR-Perf.} \\ \hline
Dataset                             & lb                    & Age $\uparrow$ & \multicolumn{3}{c}{Pose $\uparrow$} & Exp. $\uparrow$ &    &                             &                              \\
                                    &                       &                & Yaw         & Pitch     & Roll      &       &         &                             &                              \\ \hline
Baseline (FFHQ)                     & -                     & 0.252          & 15.261      & 6.369     & 2.051     & 0.379 & 0.329   & 0.999                       & 86.58                        \\ \hline
\multirow{6}{*}{\textsc{IDperturb}} & 0.9                   & 0.282          & 15.32       & 6.432     & 2.055     & 0.425 & 0.345   & 0.998                       & 87.51                        \\
                                    & 0.8                   & 0.321          & 15.435      & 6.479     & 2.058     & 0.466 & 0.357   & 0.993                       & 88.35                        \\
                                    & 0.7                   & 0.361          & 15.490      & 6.565     & 2.081     & 0.496 & 0.367   & 0.978                       & 88.73                        \\
                                    & 0.6                   & 0.402          & 15.552      & 6.628     & 2.087     & 0.519 & 0.376   & 0.946                       & 88.54                        \\
                                    & 0.5                   & 0.446          & 15.590      & 6.709     & 2.106     & 0.539 & 0.384  & 0.894                       & 88.79                        \\
                                    & 0.4                   & 0.488          & 15.655      & 6.792     & 2.134     & 0.557 & 0.391   & 0.825                       & 87.58                        \\ \hline
\end{tabular}
\label{tab:diversity_consistency_ffhq}
\end{table*}

\begin{table*}[ht!]
\centering
\caption{
Intra-class diversity and consistency of recent competitors and ours \textsc{IDperturb} on top of the Baseline(C-WF) reported as age entropy, facial expression (Exp.) entropy, head-pose STD, $D_{intra}$ and $C_{intra}$. Higher age and expression entropy, and pose STD, and  $D_{intra}$ indicate greater intra-class diversity. The first row shows authentic C-WF results.
}
\begin{tabular}{cc|cccclcll}
\hline
\multicolumn{1}{l}{}              & \multicolumn{1}{l|}{}  & \multicolumn{5}{c}{Attributes}                                                   & D-Intra & \multicolumn{1}{c}{C-Intra} & \multicolumn{1}{c}{FR-Perf.} \\ \hline
Dataset                           &                        & Age $\uparrow$ & \multicolumn{3}{c}{Pose $\uparrow$} & Exp. $\uparrow$           &         &                             &                              \\
                                  & lb                     &                & Yaw        & Pitch      & Roll      &                           &         &                             &                              \\ \hline
C-WF                              & -                      & 0.354          & 23.479     & 9.069      & 5.154     & 0.589                     & 0.423   & 0.948                       & 94.63                        \\ \hline
\multirow{6}{*}{\textsc{IDperturb}} & 0.9                    & 0.325          & 19.907     & 8.384      & 3.606     & 0.492                     & 0.393   & 0.998                       & 92.68                        \\
                                  & 0.8                    & 0.369          & 20.899     & 8.727      & 4.029     & 0.538                     & 0.417   & 0.994                       & 93.31                        \\
                                  & 0.7                    & 0.416          & 21.816     & 9.117      & 4.551     & 0.574                     & 0.439   & 0.978                       & 93.44                        \\
                                  & 0.6                    & 0.461          & 22.735     & 9.467      & 5.124     & 0.603                     & 0.458   & 0.939                       & 93.62                        \\
                                  & 0.5                    & 0.501          & 23.611     & 9.724      & 5.602     & 0.621                     & 0.474   & 0.875                       & 93.56                        \\
                                  & 0.4                    & 0.538          & 24.279     & 9.957      & 6.040     & 0.636                     & 0.487   & 0.790                       & 93.36                        \\ \hline
DCFace                            & \multicolumn{1}{l|}{-} & 0.462          & 17.926     & 7.257      & 3.059     & \multicolumn{1}{c}{0.583} & 0.368   & \multicolumn{1}{c}{0.965}   & 89.56                        \\
SFace-60                          & \multicolumn{1}{l|}{-} & 0.426          & 24.567     & 9.715      & 5.402     & \multicolumn{1}{c}{0.627} & 0.405   & \multicolumn{1}{c}{0.777}   & 77.71                        \\
DigiFace-6k                       & \multicolumn{1}{l|}{-} & 0.286          & 26.411     & 11.133     & 4.689     & \multicolumn{1}{c}{0.658} & 0.427   & \multicolumn{1}{c}{0.995}   & 83.45                        \\
UIFace                            & \multicolumn{1}{l|}{-} & 0.386          & 21.478     & 8.593      & 4.103     & \multicolumn{1}{c}{0.608} & 0.414   & \multicolumn{1}{c}{0.988}   & 93.27                        \\ \hline
\end{tabular}
\label{tab:diversity_consistency_sota}
\end{table*}

\section{Appendix C: Face recognition performance using different loss functions and network architectures}
\label{sec:app_c}

In the main submission, we followed established protocols from prior synthetic-based FR works and employed ResNet-50 \cite{He2015DeepRL,Deng_2022} as the backbone architecture, trained using the CosFace loss \cite{wang2018cosfacelargemargincosine} function. To provide a more comprehensive evaluation, we extend our experiments by training face recognition models on datasets generated with our proposed \textsc{IDperturb} (using a cosine lower bound $\mathbf{lb} = 0.6$ and classifier-free guidance scale $\omega = 1$) based on the IDiff-Face model originally trained on Casia-WebFace. We report performance across multiple training losses, including CosFace \cite{wang2018cosfacelargemargincosine}, ArcFace \cite{Deng_2022}, and AdaFace \cite{DBLP:conf/cvpr/Kim0L22}, all utilizing ResNet-50. In addition, we assess two alternative network architectures—ResNet-100 \cite{He2015DeepRL,Deng_2022} and Vision Transformer (ViT) \cite{DBLP:conf/iclr/DosovitskiyB0WZ21,DBLP:conf/cvpr/KimS0JL24}, Vit-Small, to further examine the generalizability of \textsc{IDperturb} generated data across diverse FR model designs.
The achieved results are presented in Table \ref{tab:loss_func}

\section{Appendix D: Intra-Class Diversity and Consistency of \textsc{IDperturb} on top of Baseline (FFHQ)}
\label{sec:app_d}
In the main submission, we reported the age entropy, facial-expression entropy, standard deviation of head pose, intra-class perceptual diversity, and intra-class consistency of \textsc{IDperturb} applied on top of the Baseline (C-WF). For completeness, Table~\ref{tab:diversity_consistency_ffhq} additionally reports intra-class diversity and identity consistency, together with face-recognition performance, when \textsc{IDperturb} is applied on the Baseline (FFHQ). Consistent with the observations from Table 2 of the main submission, decreasing the lower bound $lb$ from 0.9 to 0.4 results in a gradual increase in intra-class age and facial-expression variation without compromising identity coherence. The intra-class diversity $D_{intra}$ increases monotonically as $lb$ decreases, confirming that \textsc{IDperturb} effectively expands the perceptual manifold of each identity. These results further support the analysis presented in the main submission.

\section{Appendix E: Intra-Class Diversity and Consistency of \textsc{IDperturb} compared with several SOTA approaches}
\label{sec:app_e}
We report in Table \ref{tab:diversity_consistency_sota} the age entropy, facial expression entropy, STD of head-pose, intra-class perceptual diversity, and intra-class consistency.  The results are reported on our \textsc{IDperturb} ($lb \in [0.9,0.8,0.7,0.6,0.5,0.4]$) and compared with DCFace~\cite{DBLP:conf/cvpr/Kim00023} introduces dual conditioning with identity and style inputs to control appearance variation, where diversity is achieved by sampling styles from a learned style bank.
SFace~\cite{Boutros2022SFace} is a class-conditional GAN built on StyleGAN2-ADA~\cite{Karras2020StyleGANADA} without explicit variation conditions.
DigiFace~\cite{DigiFace1M} generates identities through physically based rendering using a computer graphics pipeline.
UIFace~\cite{DBLP:conf/iclr/LinHXMZD25} employs a two-phase sampling strategy, generating the early denoising steps unconditionally and introducing the identity condition only in the later steps, which promotes greater stochastic variation across samples.

Overall, \textsc{IDperturb} demonstrates a superior balance between intra-class diversity and identity consistency across all metrics. As the cosine lower bound $lb$ decreases, we observe a consistent increase in diversity indicators---age and expression entropy, head-pose STD, and $D_{\text{intra}}$, while maintaining high identity coherence ($C_{\text{intra}}$) and strong face recognition (FR) performance. This confirms that our angular perturbation strategy effectively introduces  variations in facial appearance without compromising identity coherence.

In contrast, DCFace, which achieves diversity through a style bank, exhibits lower expression entropy and pose variation. This could be attributed to a lack of specific variations in the style bank.  SFace, while showing higher variation, suffers from poor identity consistency, reflected by its significantly lower $C_{\text{intra}}$ and FR accuracy. DigiFace, based on physically-based rendering, achieves high pose diversity but lacks the realistic appearance and  semantic richness of diffusion-based approaches. UIFace, which introduces the identity condition only in the later denoising stages, attains moderate diversity with strong identity consistency, achieving closed, slightly lower, FR performance to our \textsc{IDperturb}.

\begin{table*}[ht!]
\centering
\caption{Verification accuracies of FRs trained on \textsc{IDperturb} using CosFace, ArcFace and AdaFace loss functions as well as using different backbones, ResNet50, ResNet100 and Vit-S. }
\begin{tabular}{llllllll}
Network architecture      & Loss function & \multicolumn{6}{l}{Verfication accuracies (\%)}   \\ \hline
                          &               & LFW   & AgeDB & CFP-FP & CA-LFW & CP-LFW & Avg.   \\ \hline
\multirow{3}{*}{ResNet50} & CosFace       & 99.40 & 93.20 & 93.61  & 93.50  & 88.37  & \textbf{93.62 } \\
                          & ArcFace       & 99.27 & 92.17 & 93.06  & 93.00  & 87.73  & 93.05 \\
                          & AdaFace       & 99.38 & 92.92 & 93.97  & 93.40  & 88.20  & 93.60  \\  \hline \hline
ResNet50                  & CosFace       & 99.40 & 93.20 & 93.61  & 93.50  & 88.37  & 93.62  \\
ResNet100                 & CosFace       &  99.37     &   93.69    &   94.09     &   93.57     &   89.02     &    \textbf{93.95 }   \\
Vit-Small                     & CosFace       & 99.12     &   90.71    &  90.32      &   92.70     &   86.22     &   91,81 \\   \hline
\end{tabular}
\label{tab:loss_func}
\end{table*}

\section{Appendix F: Identity-Separability synthetic-based FR}
\label{sec:app_f}
Figures \ref{fig:id_sep_ffhq}, \ref{fig:id_sep_CASIA_lb} and \ref{fig:id_sep_CASIA_w} present the genuine and impostor score distributions of the data generated by our \textsc{IDperturb} trained on different datasets (FFHQ and CASIA) for different sampling hyperparameter values ($\mathbf{lb}$ and $\omega$). These results correspond to the analysis of the identity separability metrics displayed in Tables 1 and 3 of the main submission.

\section{Appendix G: Use of existing assets and face recognition evaluation benchmarks}
\label{sec:app_g}

\subsection{Use of existing assets}
The results of the SOTA face recognition models (in the main submission) are taken directly from their corresponding works. The pretrained IDiff-Face \cite{DBLP:conf/iccv/BoutrosGKD23,DBLP:conf/iclr/LinHXMZD25}diffusion models on FFHQ and Casia-WebFace are provided under Attribution-NonCommercial-ShareAlike 4.0 
International (CC BY-NC-SA 4.0) license and Apache-2.0 license, respectively.  Pretrained IDiff-Face on FFHQ is provided under \url{https://github.com/fdbtrs/IDiff-Face} and the one trained on Casia-WebFace is provided under \url{https://github.com/Tencent/TFace/}. Both models are trained using the same architectures and training setups described in the paper \cite{DBLP:conf/iccv/BoutrosGKD23}. We provide the following details on the datasets used to train these models.
\textbf{CASIA-WebFace \cite{DBLP:conf/iclr/LinHXMZD25,DBLP:conf/cvpr/Kim0L22}:}
CASIA-Webface consists of 494,141 face images from 10,757 different identities. A prepossessed (aligned and cropped) version of CASIA-WebFace is publicly available in the InsightFace (\url{https://insightface.ai/}) repository  under Dataset-Zoo
\url{https://github.com/deepinsight/insightface/tree/master/recognition/_datasets_ }. 
The code and the databases of InsightFace are under MIT licence (\url{https://github.com/deepinsight/insightface/blob/master/LICENSE}).

\textbf{FFHQ \cite{FFHQ}}
Flickr-Faces-HQ Dataset (FFHQ) consists of 70,000 high-quality PNG images. The data is available for research under \url{https://github.com/NVlabs/ffhq-dataset}.

\subsection{Evaluation benchmarks}

This section presents the description and license information of the FR evaluation datasets used in our work.

\textbf{AgeDB-30 \cite{AgeDB30Database}:} AgeDB is an in-the-wild dataset for age-invariant face verification evaluation, containing 16,488 images of 568 identities. Every image is annotated with respect to the identity, age, and gender attribute. In our case, we report the performance for AgeDB-30 (30 years age gap) as it is the most reported and challenging subset of AgeDB. More details on the collection process can be found in \cite{AgeDB30Database} and the details on the license are presented in \url{https://ibug.doc.ic.ac.uk/resources/agedb/}.

\textbf{LFW \cite{LFWDatabase}}: Labeled Faces in the Wild (LFW) is an unconstrained face verification dataset. The LFW contains 13,233 images of 5749 identities collected from the web. The LFW is licensed under CC-BY-4.0, and more information on database creation can be found in \cite{LFWDatabase} and \url{http://vis-www.cs.umass.edu/lfw/}.

\textbf{CFP-FP \cite{CFPFPDatabase}:} Celebrities in Frontal-Profile in the Wild (CFP-FP) \cite{CFPFPDatabase} dataset addresses the comparison between frontal and profile faces. CFP-FP dataset contains 7,000 images across 500 identities, where 10 frontal and 4 profile image per identity.
More information can be found in \cite{CFPFPDatabase} and \url{http://www.cfpw.io/}.

\textbf{CALFW \cite{CALFWDatabase}:} The Cross-age LFW (CALFW) dataset \cite{CALFWDatabase} is based on LFW with a focus on comparison pairs with the age gap, however not as large as AgeDB-30. Age gap distribution of the CALFW is provided in \cite{CALFWDatabase}. It contains 3000 genuine comparisons, and the negative pairs are selected of the same gender and race to reduce the effect of attributes. The detailed information on database creation can be found in \cite{CALFWDatabase} and \url{http://whdeng.cn/CALFW/}.

\textbf{CPLFW \cite{CPLFWDatabase}:} The Cross-Pose LFW (CPLFW) dataset \cite{CPLFWDatabase} is based on LFW with a focus on comparison pairs with pose differences. CPLFW contains 3000 genuine comparisons, while the negative pairs are selected of the same gender and race. More information can be found in \cite{CPLFWDatabase} and \url{http://whdeng.cn/CPLFW/}.

\textbf{IJB-C \cite{DBLP:conf/icb/MazeADKMO0NACG18}:} The IARPA Janus Benchmark–C (IJB-C) \cite{DBLP:conf/icb/MazeADKMO0NACG18} is a video-based face recognition dataset provided by the Nation Institute for Standards and Technology (NIST). It is an extension of the IJB-B \cite{DBLP:conf/icb/MazeADKMO0NACG18} dataset with a total of 31,334 still images and 117,542 frames of 11,779 videos across 3531 identities. IJB-C is made available under different Creative Commons license variants. Detailed information on database creation can be found in \cite{DBLP:conf/icb/MazeADKMO0NACG18} and \url{https://www.nist.gov/programs-projects/face-challenges}.

\subsection{Similarity Threshold} A similarity threshold of 0.3 in Equation 8 (main submission) is selected following \cite{DBLP:conf/cvpr/Kim0L22}. This choice corresponds to the operating point of the underlying FR evaluation model, which reports a verification accuracy of 97.17\% at a threshold of 0.3080 using $TPR@FPR = 0.01\%$ on the IJB-B benchmark \cite{inproceedingsijbb}

\section*{Computing Infrastructure}
This paper utilizes a computing infrastructure comprising 2 NVIDIA RTX A6000 GPUs to conduct the experiments. The code was tested on a Linux-based operating system, using 
PyTorch version 1.13.1. The versions of all relevant software libraries and frameworks are listed in the requirements file available in the code.zip folder.

\section{Appendix H: Social Impact}
\label{sec:app_h}
This work is part of an ongoing research direction aimed at enabling the development of face recognition systems using fully synthetic data, thereby mitigating potential legal, ethical, and societal concerns associated with collecting and utilizing authentic biometric data. We emphasize that the advancement of synthetic-based face recognition serves to enhance the security, convenience, and overall quality of life for individuals—for instance, by facilitating secure access to financial or healthcare services, or improving the reliability of border control systems, all within well-defined legal and regulatory frameworks. Nonetheless, we firmly acknowledge and reject any malicious, unlawful, or discriminatory uses of this and related machine learning technologies.

{
    \small
    \bibliographystyle{ieeenat_fullname}
    \bibliography{main}
}
